\documentclass{article}
\usepackage[accepted]{icml2020}

\usepackage[utf8]{inputenc} 
\usepackage[T1]{fontenc}    
\usepackage{url}            
\usepackage{amsfonts}       
\usepackage{nicefrac}       
\usepackage{algorithm}
\usepackage{hyperref}
\usepackage{algorithmic}
\usepackage{flushend}
\usepackage{amsmath}
\date{}
\usepackage{float}        
\usepackage{subcaption}     
\usepackage{graphicx}      
\usepackage{booktabs}  
\usepackage{multirow}
\usepackage{makecell}

\usepackage{xr}

\usepackage{scrextend}

\usepackage{colortbl} 
\definecolor{hd_cl}{rgb}{0.74,0.88,0.91}
\definecolor{header_color}{rgb}{0.74,0.88,0.91}
\definecolor{ev_cl}{rgb}{0.9,0.9,0.9}
\definecolor{even_color}{rgb}{0.9,0.9,0.9}
\definecolor{subhd_cl}{rgb}{0.85,0.93,0.95}
\definecolor{subheader_color}{rgb}{0.85,0.93,0.95}
\definecolor{childheader_color}{rgb}{1.0,0.93,0.87}

\usepackage{amssymb}

\usepackage[dvipsnames]{xcolor}

\icmltitlerunning{Bayesian Optimisation over Multiple Continuous and Categorical Inputs}

\begin{document}

\twocolumn[

\icmltitle{Bayesian Optimisation over Multiple Continuous and Categorical Inputs}


\begin{icmlauthorlist}

\icmlauthor{Binxin Ru$\text{}^*$} {ox} 
\icmlauthor{Ahsan S. Alvi$\text{}^*$} {ox}
\icmlauthor{Vu Nguyen} {ox}
\icmlauthor{Michael A. Osborne } {ox}
\icmlauthor{Stephen J Roberts } {ox}

\end{icmlauthorlist}

\icmlaffiliation{ox}{University of Oxford}

\icmlcorrespondingauthor{Binxin Ru}{robin@robots.ox.ac.uk}



\icmlkeywords{Machine Learning, ICML}

\vskip 0.3in
]

\printAffiliationsAndNotice{\icmlEqualContribution} 

\begin{abstract}

Efficient optimisation of black-box problems that comprise both continuous and categorical inputs is important, yet poses significant challenges. Current approaches, like one-hot encoding, severely increase the dimension of the search space, while separate modelling of category-specific data is sample-inefficient. Both frameworks are not scalable to practical applications involving \emph{multiple} categorical variables, each with \emph{multiple} possible values. We propose a new approach, Continuous and Categorical Bayesian Optimisation (CoCaBO), which combines the strengths of multi-armed bandits and Bayesian optimisation to select values for both categorical and continuous inputs. We model this mixed-type space using a Gaussian Process kernel, designed to allow sharing of information across \emph{multiple} categorical variables; this allows CoCaBO to leverage all available data efficiently. We extend our method to the batch setting and propose an efficient selection procedure that dynamically balances exploration and exploitation whilst encouraging batch diversity. 
We demonstrate empirically that our method outperforms existing approaches on both synthetic and real-world optimisation tasks with continuous and categorical inputs.

\end{abstract}

\section{Introduction}
Existing work has shown Bayesian Optimisation (BO) to be remarkably successful at optimising functions with continuous input spaces \cite{Snoek_2012Practical, Hennig_2012Entropy, Hernandez_2015Predictive, ru2018fast, Shahriari_2016Taking, frazier2018tutorial, alvi2019asynchronous}. 
However, in many situations, optimisation problems involve a mixture of continuous and categorical variables. 
For example, with a deep neural network, we may want to adjust the learning rate and the weight decay (continuous), as well as the activation function type in each layer (categorical). 
Similarly, in a gradient boosting ensemble of decision trees, we may wish to adjust the learning rate and the maximum depth of the trees (continuous), as well as the boosting algorithm and loss function (categorical).

Having a mixture of categorical and continuous variables presents challenges. 
If some inputs are categorical variables, then the common assumption that the BO acquisition function is differentiable over the input space, which allows the acquisition function to be efficiently optimised, is no longer valid.
Recent research has dealt with categorical variables in different ways. The simplest approach for BO with Gaussian process (GP) \cite{Rasmussen_2006gaussian} surrogates is to use a one-hot encoding on the categorical variables, so that they can be treated as continuous variables, and perform BO on the transformed space \cite{gpyopt2016}. Alternatively,
the mixed-type inputs can be handled using a hierarchical structure, such as using random forests \cite{Hutter_2011Sequential,Bergstra_2011Algorithms} or multi-armed bandits (MABs) \cite{gopakumar2018algorithmic_NIPS}.
These approaches come with their own challenges, which we will discuss below (see Section \ref{sec:related}). 
In particular, the existing approaches are not well designed for \emph{multiple} categorical variables with \emph{multiple} possible values. 
Additionally, no GP-based BO methods have explicitly considered the batch setting for continuous-categorical inputs, to the best of our knowledge.

In this paper, we present a new Bayesian optimisation approach for optimising a black-box function with multiple continuous and categorical inputs, termed Continuous and Categorical Bayesian Optimisation (CoCaBO). Our approach is motivated by the success of MABs \cite{auer2002finite,auer2002nonstochastic} in identifying the best value(s) from a discrete set of options. 

Our main contributions are as follows:
\begin{itemize}
    \item We propose a method  which combines the strengths of MABs and BO to optimise black-box functions with \emph{multiple} categorical  and continuous inputs. (Section \ref{sec:cocabo}). 
    \item We present a GP kernel to capture complex interactions between the continuous and categorical inputs (Section \ref{ssec:surrogate}). Our kernel allows sharing of information across different categories without resorting to one-hot transformations.
    \item We introduce a novel batch selection method for mixed input types that extends CoCaBO to the parallel setting, and dynamically balances exploration and exploitation and encourages batch diversity (Section \ref{ssec:batch_selection}). 
    \item We demonstrate the effectiveness of our methods on a variety of synthetic and real-world optimisation tasks with \emph{multiple} categorical and continuous inputs (Section \ref{sec:experiments}).
 
\end{itemize}

\section{Preliminaries}

\label{sec:prelim}


In this paper, we consider the problem of optimising a black-box function $f(\mathbf{z})$ where the input $\mathbf{z}$ consists of both continuous and categorical inputs, $\mathbf{z}=[\mathbf{h},\mathbf{x}]$, where $\mathbf{h} = [h_1, \ldots, h_c]$ are the categorical variables, with each variable $h_j \in \{1,2,\ldots,N_j\}$ taking one of $N_j$ different values, and $\mathbf{x}$ is a point in a  $d$-dimensional hypercube $\mathcal{X}$. 
Formally, we aim to find the best configuration to maximise the black-box function
\begin{equation}
\label{eq:optimisation}
    \mathbf{z}^*=[\mathbf{h}^*, \mathbf{x}^*] = \arg\max_\mathbf{z} f(\mathbf{z})
\end{equation}
by making a series of evaluations $\mathbf{z}_1$,..., $\mathbf{z}_T$. 
Later we extend our method to allow parallel evaluation of multiple points, by selecting a batch $\{ \mathbf{z}_t^{(i)} \}_{i=1}^b$ at each optimisation step $t$. 

Bayesian optimisation \cite{Brochu_2010Tutorial, Shahriari_2016Taking,nguyen2019knowing} is an approach for optimising a black-box function ${\mathbf{x}^{*}=\arg\max_{\mathbf{x}\in\mathcal{X}}f(\mathbf{x})}$ such that its optimal value is found using a small number of evaluations. 
BO often uses a Gaussian process \cite{Rasmussen_2006gaussian} surrogate to model the objective $f$.
A GP defines a probability distribution over functions $f$,
as $f(\mathbf{x})\sim \texttt{GP}\left(m\left(\mathbf{x}\right),k\left(\mathbf{x},\mathbf{x}'\right)\right)$,
where $m\left(\mathbf{x}\right)$ and $k\left(\mathbf{x},\mathbf{x}'\right)$ are the mean
and covariance functions respectively, which encode our prior beliefs about $f$.
Using the GP posterior, BO defines an acquisition function $\alpha_{t}\left(\mathbf{x}\right)$ which is optimised to identify the next location to sample $\mathbf{x}_{t}=\arg\max_{x\in\mathcal{X}}\alpha_{t}\left(\mathbf{x}\right)$.
Unlike the original objective function $f(\mathbf{x})$, the acquisition function $\alpha_t\left(\mathbf{x}\right)$ is cheap to compute and can be optimised using standard techniques.






\section{Related Work}
\label{sec:related}

\subsection{One-hot encoding}
\label{ssec:one-hot}
A common method for dealing with categorical variables is to transform them into a one-hot encoded representation, where a variable with $N$ choices is transformed into a vector of length $N$ with a single non-zero element. 
This is the approach followed by popular BO packages like Spearmint \cite{Snoek_2012Practical} and GPyOpt \cite{Gonzalez_2015Batch, gpyopt2016}. 

There are two main drawbacks with this approach. 
First, the commonly-used kernels in the GP surrogate assume that $f$ is continuous and differentiable in the input space, which is not the case for one-hot encoded variables, as the objective is only defined for a small subspace within this representation.


The second drawback is that one-hot encoding adds many dimensions to the search space, making the continuous optimisation of the acquisition function much harder. Additionally, the distances between one-hot encoded categorical variables tend to be large. As a result, the optimisation landscape is characterised by many flat regions, increasing the difficulty of optimisation \cite{Rana_ICML2017High}.

\subsection{Category-specific continuous inputs}\label{subsec:exp3bo}
We consider two recent approaches in handling category-specific continuous inputs, a specific setting of mixed categorical-continuous.
The first approach is EXP3BO \cite{gopakumar2018algorithmic_NIPS}, which can deal with mixed categorical and continuous input spaces by utilising a MAB. When the categorical variable is selected by the MAB, EXP3BO constructs a GP surrogate specific to the chosen category for modelling the continuous domain, i.e. it shares no information across the different categories. 
The observed data are divided into smaller subsets, one for each category, and as a result EXP3BO can handle only a small number of categorical choices and requires a large number of samples. Recently, \citet{nguyen2019bayesian} considered extending the category-specific continuous inputs to the batch setting using Thompson sampling \cite{russo2018tutorial}.

\subsection{Hierarchical approaches}
\label{ssec:hierarchical}
Random forests (RFs) \cite{Breiman_2001Random} can naturally consider continuous and categorical variables, and are used in  SMAC  \cite{Hutter_2011Sequential} as the underlying surrogate model for $f$. 
However, the predictive distribution of the RF, which is used to select the next evaluation, is less reliable, as it relies on randomness introduced by the bootstrap samples and the randomly chosen subset of variables to be tested at each node to split the data.
Moreover, RFs can easily overfit and we need to carefully choose the number of trees. 
Another tree-based approach is Tree Parzen Estimator (TPE) \cite{Bergstra_2011Algorithms}, an optimisation algorithm based on tree-structured Parzen density estimators. 
TPE uses nonparametric Parzen kernel density estimators to model the distribution of good and bad configurations w.r.t. a reference value. Due to the nature of kernel density estimators, TPE also supports continuous
and discrete spaces.

\subsection{Integer-continuous inputs}
There are also other approaches for integer-continuous variables \cite{swiler2014surrogate,garrido2019dealing, daxberger2019mixed,pelamatti2020overview} which are related, but not essentially suitable for mixed categorical-continuous variables.

\paragraph{Remark.} To the best of our knowledge, none of the above approaches has solved satisfactorily for the mixed type variables where (1) the continuous variables are not specific to a category and  (2) we have \emph{multiple} categorical variables each of which include \emph{multiple} choices.

\section{Continuous and Categorical Bayesian Optimisation (CoCaBO)}

\label{sec:cocabo}


\begin{algorithm}[t]
	    \caption{CoCaBO Algorithm }\label{alg:CoCaBO_sequential}
	\begin{algorithmic}[1]
		\vspace{0.5em}
		\STATE {\bfseries Input:}  A black-box function $f$, observation data $\mathcal{D}_0$, maximum number of iterations $T$
		\STATE {\bfseries Output:} The best recommendation $\mathbf{z}_T=[\mathbf{x}_T, \mathbf{h}_T]$
        \FOR{$t=1, \dots, T$}
            \STATE Select $\mathbf{h}_t = [ h_{1,t}, \ldots , h_{c,t}] \leftarrow$ MAB($\{ \mathbf{h}_i, f_i\}_i^{t-1}$)
        	\STATE Select $\mathbf{x}_{t} = \arg\max \alpha_t (\mathbf{x} \vert \mathcal{D}_{t-1}, \mathbf{h}_t)$
        	\STATE Query at $\mathbf{z}_{t}=[\mathbf{x}_t, \mathbf{h}_t]$ to obtain $f_{t}$
        	\STATE $\mathcal{D}_{t} \leftarrow \mathcal{D}_{t-1} \cup (\mathbf{z}_t , f_{t})$ and update the reward distributions of MAB with ($\mathbf{h}_t , f_{t}$)
    	\ENDFOR
	\end{algorithmic}
\end{algorithm}

\begin{figure} [t] 
    \vspace{-5pt}

    \centering
	\includegraphics[trim=0cm 0cm 0cm 0cm, clip, width=0.75\linewidth]{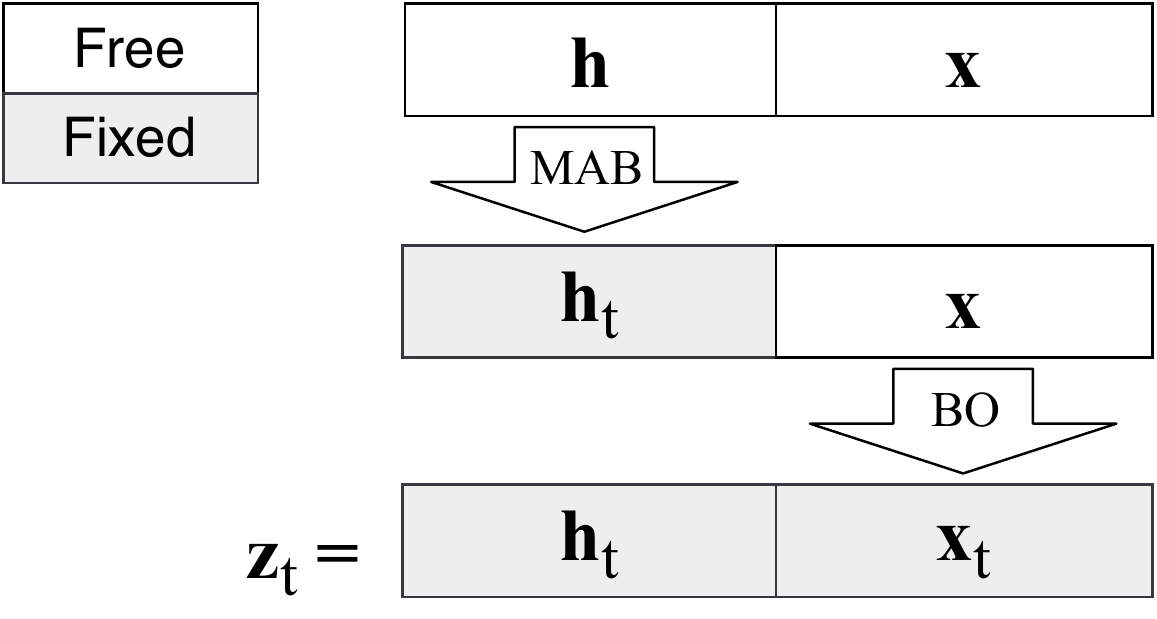}
	\caption{Optimisation procedure in CoCaBO.}
	\label{fig:cat_cont_var_selection}
	\vspace{-15pt}

\end{figure}

Our proposed method, Continuous and Categorical Bayesian Optimisation, harnesses both the advantages of multi-armed bandits to select categorical inputs and the strengths of GP-based BO in optimising continuous input spaces. The CoCaBO procedure is shown in Algorithm \ref{alg:CoCaBO_sequential}.
CoCaBO first decides the values of the categorical inputs $\mathbf{h}_t$ by using MAB (Step 4 in Algorithm \ref{alg:CoCaBO_sequential}). Given $\mathbf{h}_t$, it then maximises the acquisition function to select the continuous part $\mathbf{x}_t$ which forms the next point $\mathbf{z}_t=[\mathbf{h}_t, \mathbf{x}_t]$ for evaluation, as illustrated in Figure \ref{fig:cat_cont_var_selection}.

\subsection{CoCaBO bandit algorithm}

The design of the MAB system in the CoCaBO algorithm is motivated by two considerations. 

First, our problem setting deals with \textit{multiple} categorical inputs $\mathbf{h} = [h_1, \ldots, h_c]$ and the value of each $h_j$ is determined by a MAB agent. This results in a multi-agent MAB system and requires coordination. We achieved the coordination across the agents by using a joint reward function. Specifically, at each iteration, each agent decides the categorical value for its own variable among the corresponding $N_j$ choices but the objective function is evaluated based on the joint decision of all the agents. This function value is then used to update the reward distribution for the chosen category among all the agents. A concurrent work \cite{carlucci2020manas} also uses the similar coordination in deploying multi-agent learning system and shows its effectiveness on their neural architecture search application.

Second, we adopt the EXP3 \cite{auer2002nonstochastic} method as the MAB algorithm because it makes comparatively fewer assumptions on reward distributions and can deal with the general setting of adversarially-defined rewards. The general adversarial bandit setting is important for our application as it enables the MAB part of our algorithm to handle the non-stationary nature of the rewards observed during the optimisation \cite{allesiardo2017non}; the function value for the given values of categorical variables will improve over iterations as we apply BO on the continuous space.
In contrast, Thompson Sampling \cite{thompson1933likelihood}, UCB and $\epsilon$-greedy assume that the rewards are i.i.d. samples from stationary distributions \cite{allesiardo2017non}, thus are not suitable for our application. 

The resultant multi-agent EXP3 algorithm has a cumulative regret of $\mathcal{O}\left(c \sqrt{TN \log(N)}\right)$, which is sublinear in $T$ and increases with the number of categorical inputs $c$ as well as the number of choices $N$ for each categorical input, as derived in Appendix C. By using the MAB to decide the values for categorical inputs, we only need to optimise the acquisition function over the continuous subspace $\mathcal{X} \in \mathbb{R}^d$. In comparison to one-hot based methods, whose acquisition functions are defined over 
$\mathbb{R}^{(d+\sum_{i}^c N_i)}$, our approach enjoys a significant reduction in the difficulty and cost of optimising the acquisition function\footnote{To optimise the acquisition function to within $\zeta$ accuracy using a grid search or branch-and-bound optimiser, our approach requires only $\mathcal{O}(\zeta^{-d})$ calls and one-hot approaches require $\mathcal{O}(\zeta^{-(d+\sum_{i}^c N_i)})$ calls. 
The cost-saving grows exponentially with the number of categories $c$ and number of choices for each category $N_i$.}. 

\begin{figure*}[t]
    \centering
    \includegraphics[width=1\linewidth]{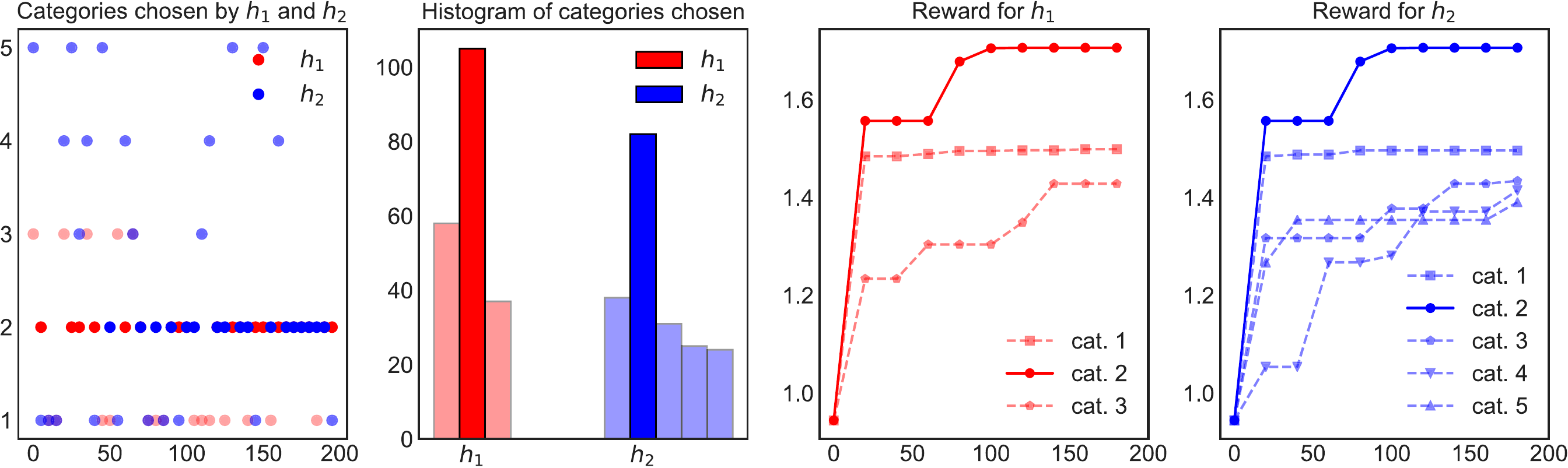}
	\caption{CoCaBO correctly optimises the two categorical inputs $h_1$ (Red) and $h_2$ (Blue) of the \textit{Func-2C} test function over 200 iterations. The \emph{best} category is $h_i = 2$ for both $h_1$ $(N_1 = 3)$ and $h_2$ $(N_2 = 5)$, and
	is highlighted in all plots. The left subplot shows the selections made by CoCaBO, showing how the both categorical inputs increasingly focus on the best categories as the algorithm progresses. The second left subplot shows the histogram of categories selected, with the best category being chosen the most frequently. The right subplots show the reward for each categorical value for $h_1$ and $h_2$ across iterations. Again, we see the correct category being identified for both categorical inputs for the highest rewards.}
	\label{fig:CoCaBO_func2C_demo}
\end{figure*}

In Figure \ref{fig:CoCaBO_func2C_demo}, we demonstrate the effectiveness of our approach in dealing with categorical variables via a simple synthetic example \textit{Func-2C} (described in Section \ref{sec:experiments}), which comprises two categorical inputs, $h_1$ ($N_1=3$) and $h_2$ ($N_2=5$), and two continuous inputs. 
The optimal function value lies in the subspace when both categorical variables  $h_1=h_2=2$. 
The categories chosen by CoCaBO at each iteration, the histogram of all selections and the rewards for each category are shown for 200 iterations. 
We can see that CoCaBO successfully identifies and focuses on the correct categories.

\subsection{CoCaBO kernel design} 
\label{ssec:surrogate}

We  propose to use a combination of two separate kernels: $k_z(\mathbf{z},\mathbf{z}')$ will combine $k_h(\mathbf{h},\mathbf{h}')$, a kernel defined over the categorical inputs, with $k_x(\mathbf{x},\mathbf{x}')$, for the continuous inputs.

For the categorical kernel, we propose using an indicator-based similarity metric, ${k_h(\mathbf{h}, \mathbf{h}') =  \frac{\sigma}{c} \sum_{i=1}^{c}\mathbb{I}(h_i - h_i')}$,
where $\sigma$ is the kernel variance and $\mathbb{I}(h_i, h_i') = 1$ if $h_i = h_i'$ and is zero otherwise. This kernel can be derived as a special case of a RBF kernel, which is explored in Appendix D.

There are several ways of combining kernels that result in valid kernels \cite{duvenaud2013structure}. One approach is to sum them together. Using a sum of kernels, that are each defined over different subsets of an input space, has been used successfully for BO in the context of high-dimensional optimisation in the past \cite{kandasamy2015high}. 
Simply adding the continuous kernel to the categorical kernel $k_x(\mathbf{x},\mathbf{x}') + k_h(\mathbf{h},\mathbf{h}')$, though, provides limited expressiveness, as this translates in practice to learning a single common trend over $\mathbf{x}$, and an offset depending on $\mathbf{h}$.

An alternative approach is to use the product of the two kernels $k_x(\mathbf{x},\mathbf{x}') \times k_h(\mathbf{h},\mathbf{h}')$. 
This form allows the kernel to encode couplings between the continuous and categorical domains, allowing a richer set of relationships to be captured,
but if there are no overlapping categories in the data, which is likely to occur in early iterations of BO, this would cause the product kernel to be zero and prevent the model from learning. 

We therefore propose the CoCaBO kernel as a mixture of the sum and product kernels:
\begin{align}
\label{eq:CoCaBO-kernel}
    k_z(\mathbf{z}, \mathbf{z}') = &   (1-\lambda)\big( k_h(\mathbf{h}, \mathbf{h}') + k_x(\mathbf{x}, \mathbf{x}') \big) \nonumber\\
    & + \lambda  k_h(\mathbf{h}, \mathbf{h}')k_x(\mathbf{x}, \mathbf{x}'),
\end{align}
The trade-off between them is controlled by a parameter $\lambda \in [0, 1]$, which can be optimised jointly with the GP hyperparameters (see Appendix E).

It is worth highlighting a key benefit of our formulation over alternative hierarchical methods discussed in Section \ref{ssec:hierarchical}. Rather than dividing our data into a subset for each combination of categories, we instead use all of our acquired data at every stage of the optimisation, as our kernel is able to combine information from data within the same category as well as from different categories, which improves its modelling performance (Section \ref{ssec:sequential_performance}). 

\begin{algorithm}[t]
	    \caption{CoCaBO batch selection}\label{alg:CoCaBO_batch_selection}
	\begin{algorithmic}[1]
		\vspace{0.5em}
		\STATE {\bfseries Input:} Observation data $\mathcal{D}_{t-1}$
		\STATE {\bfseries Output:} The batch $\mathcal{B}_t = \{\mathbf{z}_t^{(1)}, \ldots, \mathbf{z}_t^{(b)}\}$
		\STATE  $\mathbf{H}_t = \{\mathbf{h}_t^{(1)}, \ldots, \mathbf{h}_t^{(b)}\} \leftarrow \text{Batch MAB}(\mathcal{D}_{t-1})$
		\STATE $(\mathbf{u}_1, v_1), \ldots, (\mathbf{u}_q, v_q)$ are the unique categorical values in $\mathbf{H}_t$ and their counts
		\STATE Initialise $\mathcal{B}_t = \varnothing$ and $\mathcal{D}'_{t-1}=\mathcal{D}_{t-1}$
        \FOR{$j=1, \ldots, q$}
            \STATE $\{\mathbf{x}_i\}_{i=1}^{v_j} \leftarrow \text{KB}(\mathbf{u}_j, \mathcal{D}'_{t-1})$
        	\STATE $\mathbf{Z}_j = \{\mathbf{u}_j, \mathbf{x}_i\}_{i=1}^{v_j}$ and $\mathcal{B}_t \leftarrow \mathcal{B}_t \cup \mathbf{Z}_j$ 
        	\STATE $\mathcal{D}'_t \leftarrow \mathcal{D}'_{t-1} \cup \{\mathbf{Z}_j, \mu(\mathbf{Z}_j)\}$ 
    	\ENDFOR
    	\STATE {\bfseries Output:} $\mathcal{B}_t$
	\end{algorithmic}
\end{algorithm}

\begin{figure}[t]
    \centering
    \includegraphics[trim=5cm 6cm 9cm 3cm, clip, width=1.0\linewidth]{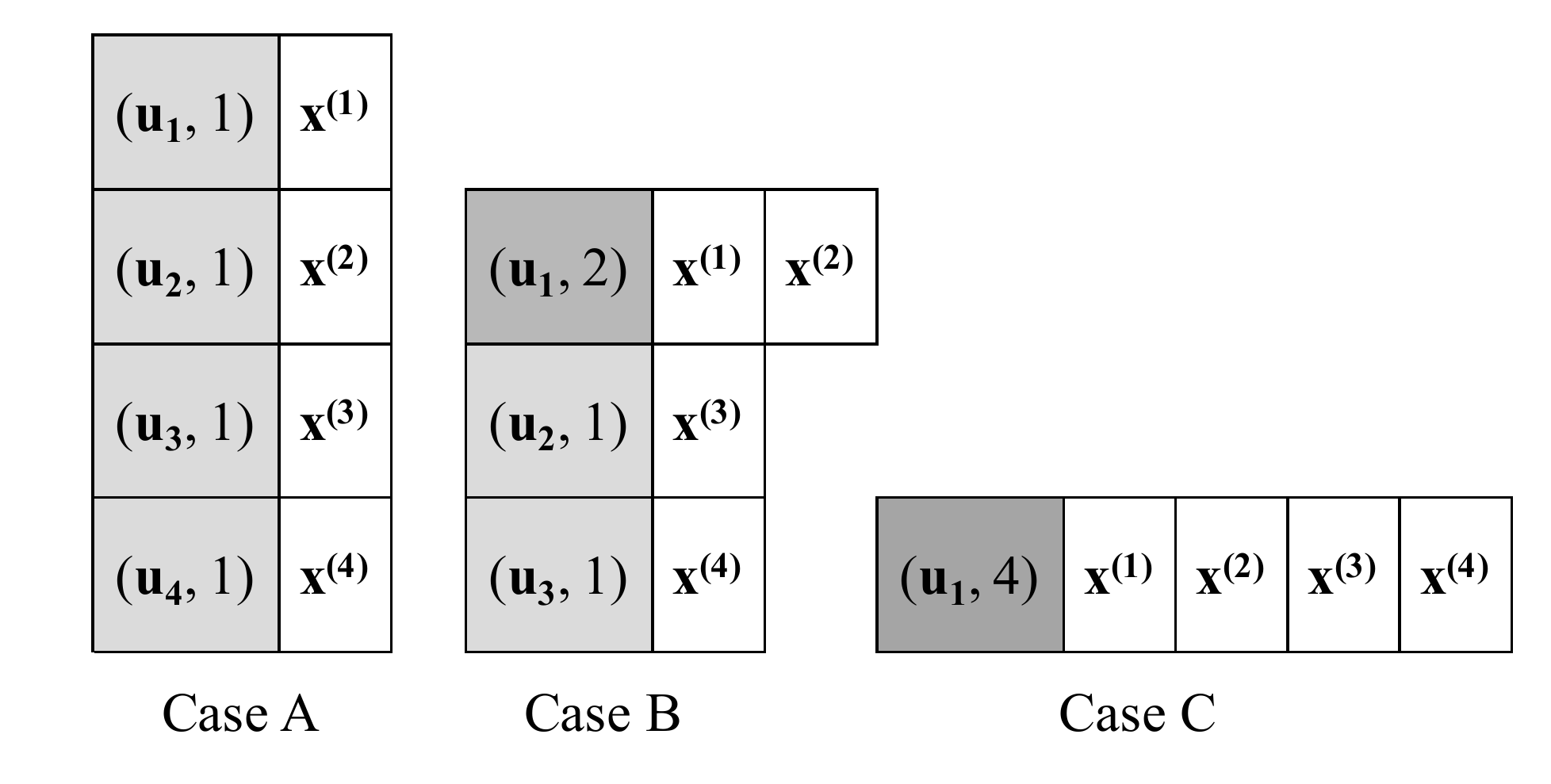}
	\caption{Three example cases for selecting a batch of 4 unique points ($b=4$). $\mathbf{u}_j$ represents a unique categorical point. If all the categorical points in the batch are different, we select 1 continuous location $\mathbf{x}^{(i)}$ conditioned on each $\mathbf{u}_j$ (Case A). If 2 out of a batch of 4 categorical points are equal to $\mathbf{u_1}$, we sequentially select 2 continuous locations $\mathbf{x}^{(1)}$ and $\mathbf{x}^{(2)}$ conditioned on $\mathbf{u_1}$ using KB and collect only 1 continuous location for the other two categorical data $\mathbf{u_2}$ and $\mathbf{u_3}$ (Case B). }
	\label{fig:batch_selection_cases}
\end{figure}



Our kernel is able to learn the covariances $k_z$ between observations in the joint input space $\mathcal{Z} = \mathcal{H} \times \mathcal{X}$. When two data points come from different categories, $k_z$ is dominated by $k_x$ (in the additive term). When there is some overlap in categorical choices between data points, then $k_z$ is determined by contributions from both $k_x$ and $k_h$ (in both additive and multiplicative terms). In comparison, the one-hot encoded kernel converts all values of the categorical variables into additional continuous dimensions, allowing it to give nonzero covariances between points from different categories. We compare the regression performance of the CoCaBO kernel and a one-hot encoded kernel on some synthetic functions in Section \ref{ssec:surrogate-regression} and show that our kernel leads to a superior predictive performance over one-hot kernel.

\subsection{Batch CoCaBO}
\label{ssec:batch_selection}

Our focus on optimising computer simulations and modelling pipelines provides a strong motivation to extend CoCaBO to select and evaluate multiple tasks at each iteration, in order to better utilise available hardware resources \cite{Snoek_2012Practical, Wu2016ThePK, Shah_2015Parallel, Contal_2013Parallel}.

The batch CoCaBO algorithm uses the ``multiple plays'' formulation of EXP3, called EXP3.M \cite{auer2002nonstochastic}, which returns a batch of categorical choices, and combines it with the Kriging Believer (KB)\footnote{Note that our approach can easily utilise other batch selection techniques if desired.} \cite{Ginsbourger_2010Kriging} batch method to select the batch points in the continuous domain.
We choose KB for the batch creation, as it can consider already-selected batch points, including those with different categorical values, without making restrictive assumptions as do other popular techniques, e.g. local penalisation \cite{Gonzalez_2015Batch, alvi2019asynchronous} assumes that $f$ is Lipschitz continuous. A detailed description of KB algorithm is in Appendix F. Our novel contribution is a method for combining the batch points selected by EXP3.M with batch BO procedures for continuous input spaces.
Assume we are selecting a batch of $b$ points ${\mathcal{B}_t = \{ \mathbf{z}_t^{(i)} \}_{i=1}^b}$ at iteration $t$. A simple approach is to select a batch of categorical variables $\mathbf{H}_t = \{ \mathbf{h}_t^{(i)}\}_{i=1}^b$ and then choose a corresponding continuous variable for each categorical point as in the sequential algorithm above, thus forming $\{ \mathbf{z}_t ^{(i)}\}_{i=1}^b = \{ \mathbf{h}_t^{(i)}, \mathbf{x}_t^{(i)} \}_{i=1}^b$. 
However, such a batch method may not identify $b$ unique locations, as some values in $\{ \mathbf{h}_t^{(i)}\}_{i=1}^b$ may be repeated and you will get identical $\mathbf{x}$ values for repeated $\mathbf{h}_t^{(i)}$ values. This is even more problematic when the number of possible combinations for the categorical variables, $\prod_{i=1}^c N_i$, is smaller than the batch size $b$, as we would never identify a full batch of unique points. 

\begin{table*}
  \caption{Categorical and continuous inputs to be optimised for real-world tasks. $N_i$ in the parentheses indicate the number of categorical choices that each categorical input has.}
  \vspace{8pt}
  \label{table:real_problem_inputs}
    \centering
    \begin{tabular}{llll}
    \toprule
     & \textit{SVM-Boston} ($d=3$, $c=3$) & \textit{XG-MNIST}($d=5$, $c=3$) & \textit{NAS-CIFAR10} ($d=22$, $c=5$) \\ \midrule
    \multirow{3}{*}{{\large $\mathbf{h}$}} 
    & kernel type ($N_1=4$),        & booster type ($N_1=2$),   & Operation choice for the 5 intermediate \\ 
    &kernel coefficient ($N_2=2$),  & grow policies ($N_2=2$),    &  node in the directed acyclic graph (DAG)\\ 
    &using shrinking ($N_3=2$)      & training objectives ($N_3=2$) & of architecture design ($\{ N_i=3\}^{5}_i$)\\
    \midrule
    \multirow{3}{*}{{\large $\mathbf{x}$}} 
    & penalty parameter,       & learning rate, regularisation, & creation probability for each of the 21 \\ 
    & tolerance for stopping,   & maximum depth, subsample,     & possible edges in the DAG, \\ 
    & model complexity   & minimum split loss  & number of edges present in the DAG \\
    \bottomrule
\end{tabular}
\end{table*}

\begin{table*}
\centering
\caption{Mean and standard error of the predictive log likelihood of the CoCaBO and the One-hot BO surrogates on synthetic test functions. Both models were trained on 250 samples and evaluated on 100 test points. The CoCaBO surrogate can model the function surface better than the One-hot surrogate as the number of categorical variables increases.}
\begin{tabular}{lcccccc}
    \toprule
       &\textit{Ackley-2C}       &\textit{Ackley-3C}      &\textit{Ackley-4C}      &\textit{Ackley-5C}      &\textit{Func-2C}       &\textit{Func-3C}      \\\cmidrule(r){2-7}
CoCaBO  & $-69.1(\pm 9.47)$       & $\bf{-74.9(\pm 3.17)}$ & $\bf{-92.0(\pm 6.31)}$ & $\bf{-114(\pm 8.31)}$ & $\bf{115(\pm 14.1)}$ & $\bf{167(\pm 6.54)}$      \\
One-hot & $\bf{-60.4(\pm 5.91)}$  & $-107(\pm 23.7)$       & $-102(\pm 7.94)$       & $-120(\pm 10.3)$      & $12.4(\pm 10.7)$     & $-81.0(\pm 4.69)$            \\    
\bottomrule
\end{tabular}
\label{tab:cocabo-vs-onehot-surrogate}
\end{table*}

Our batch selection method, outlined in Algorithm \ref{alg:CoCaBO_batch_selection}, allows us to create a batch of unique points ${\mathcal{B}_t = \{ \mathbf{z}_t^{(i)} \}_{i=1}^b}$ (i.e. $\mathbf{z}_t^{(i)} \neq \mathbf{z}_t^{(j)}$ if $i \neq j$) by allocating multiple continuous batch points to more desirable categories. The key idea is to first collect the $q$ unique categorical points \(\{\mathbf{u}_j\}_{j=1}^q\) and how often they occur \(\{v_j\}_{j=1}^q\) from the batch \(\mathbf{H}_t\).  The count $v_j$ defines how many continuous batch points will be selected via KB for each unique categorical point \(\mathbf{u}_j\). This process is illustrated in Figure \ref{fig:batch_selection_cases} for three possible scenarios. The benefit of using KB here is that the algorithm can take into account selections across the different $\mathbf{h}$ to impose diversity in the batch in a consistent manner.

\section{Experiments}
\label{sec:experiments}

\begin{figure*} [t]  
    \begin{subfigure}{0.33\linewidth}
     \centering
    \includegraphics[trim=0cm 0.cm 0cm  0.cm, clip, width=1.0\linewidth]{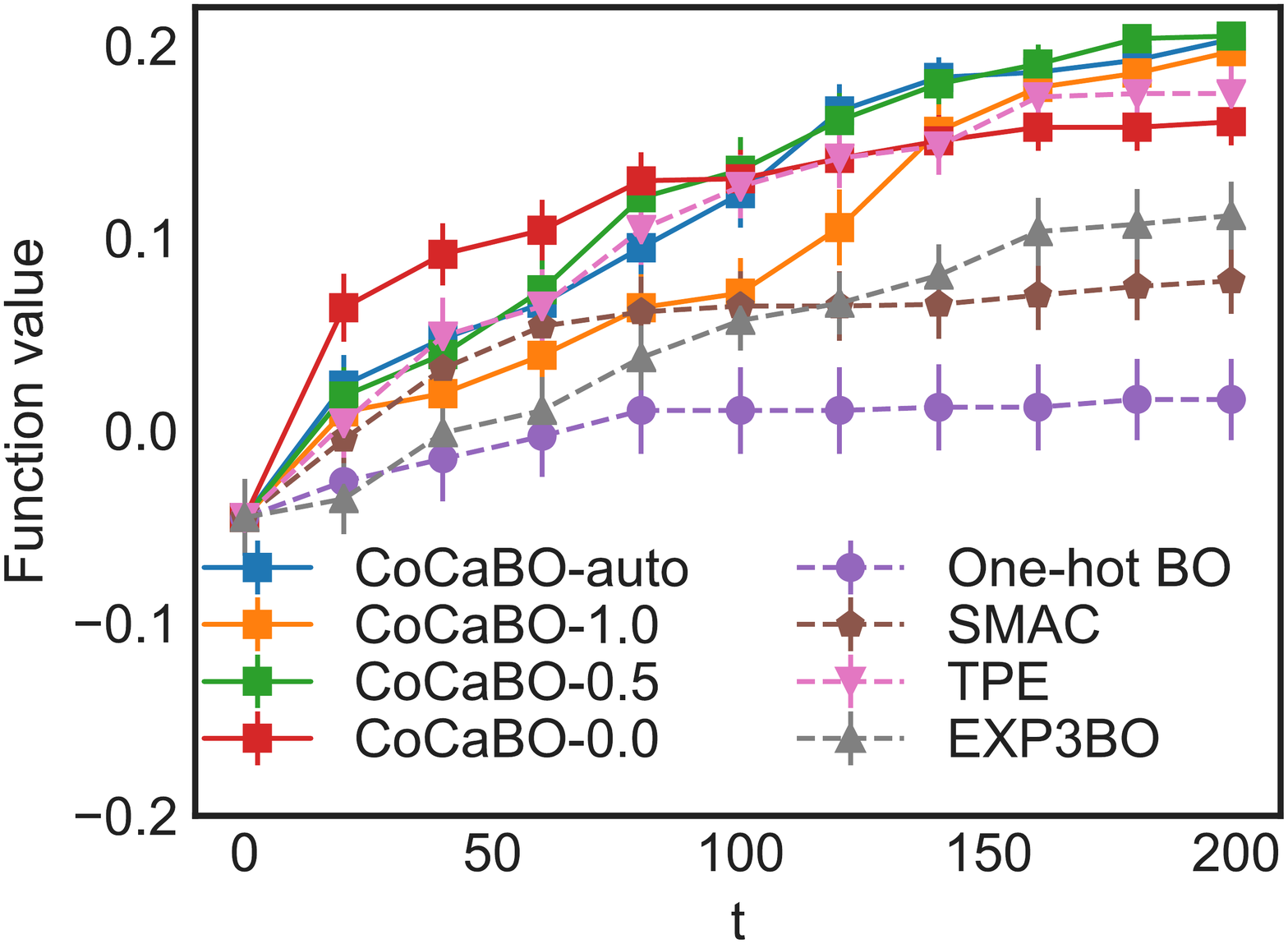}
    \caption{\textit{Func-2C}}
    \label{subfig:func2c}
    \end{subfigure}
    \begin{subfigure}{0.33\linewidth}
    \centering
    \includegraphics[trim=0cm 0.cm 0cm  0.cm, clip, width=1.0\linewidth]{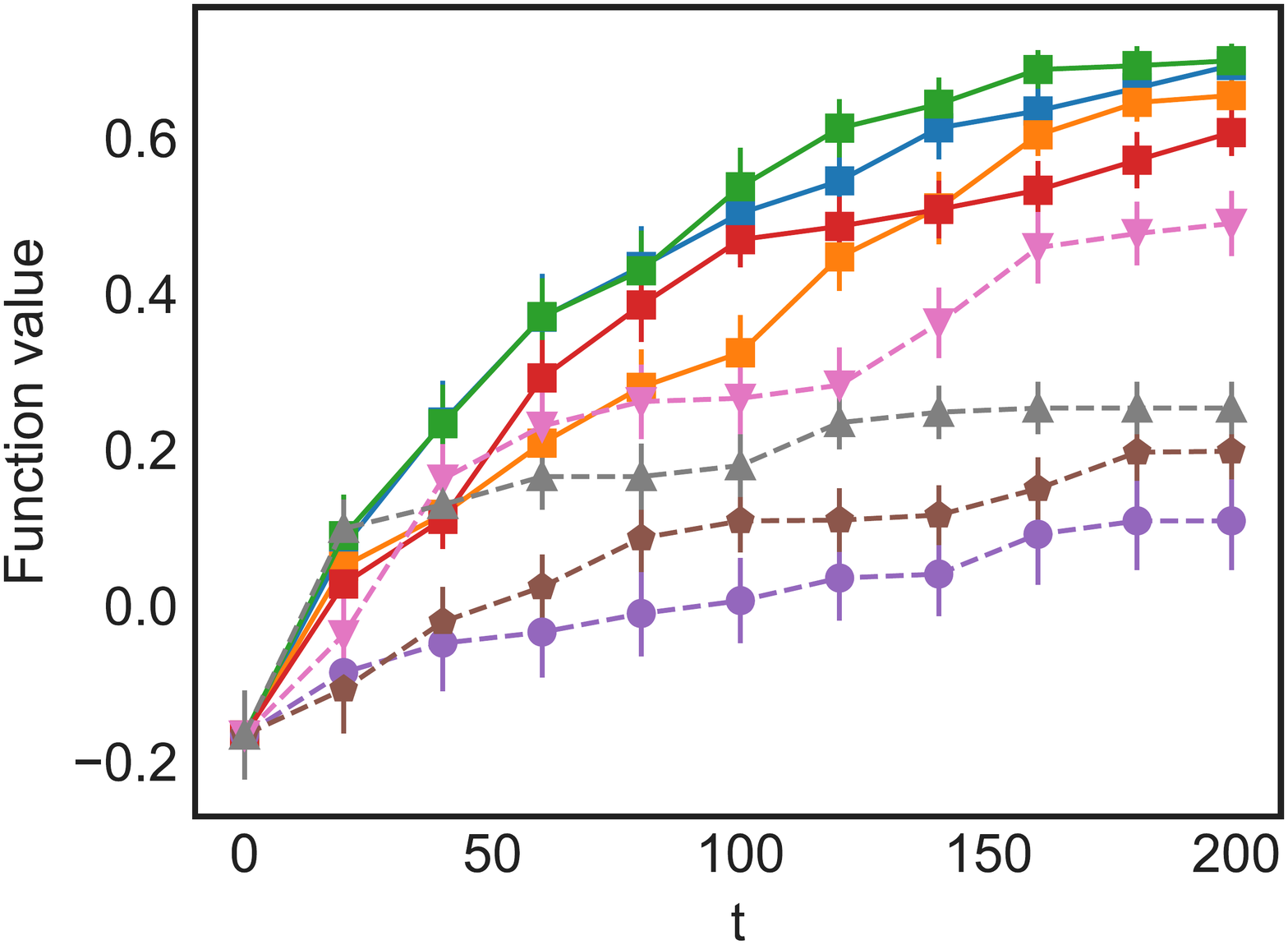}
    \caption{\textit{Func-3C}}
    \label{subfig:func3c}
    \end{subfigure}
    \begin{subfigure}{0.33\linewidth}
     \centering
    \includegraphics[trim=0cm 0.cm 0cm  0.cm, clip, width=1.0\linewidth]{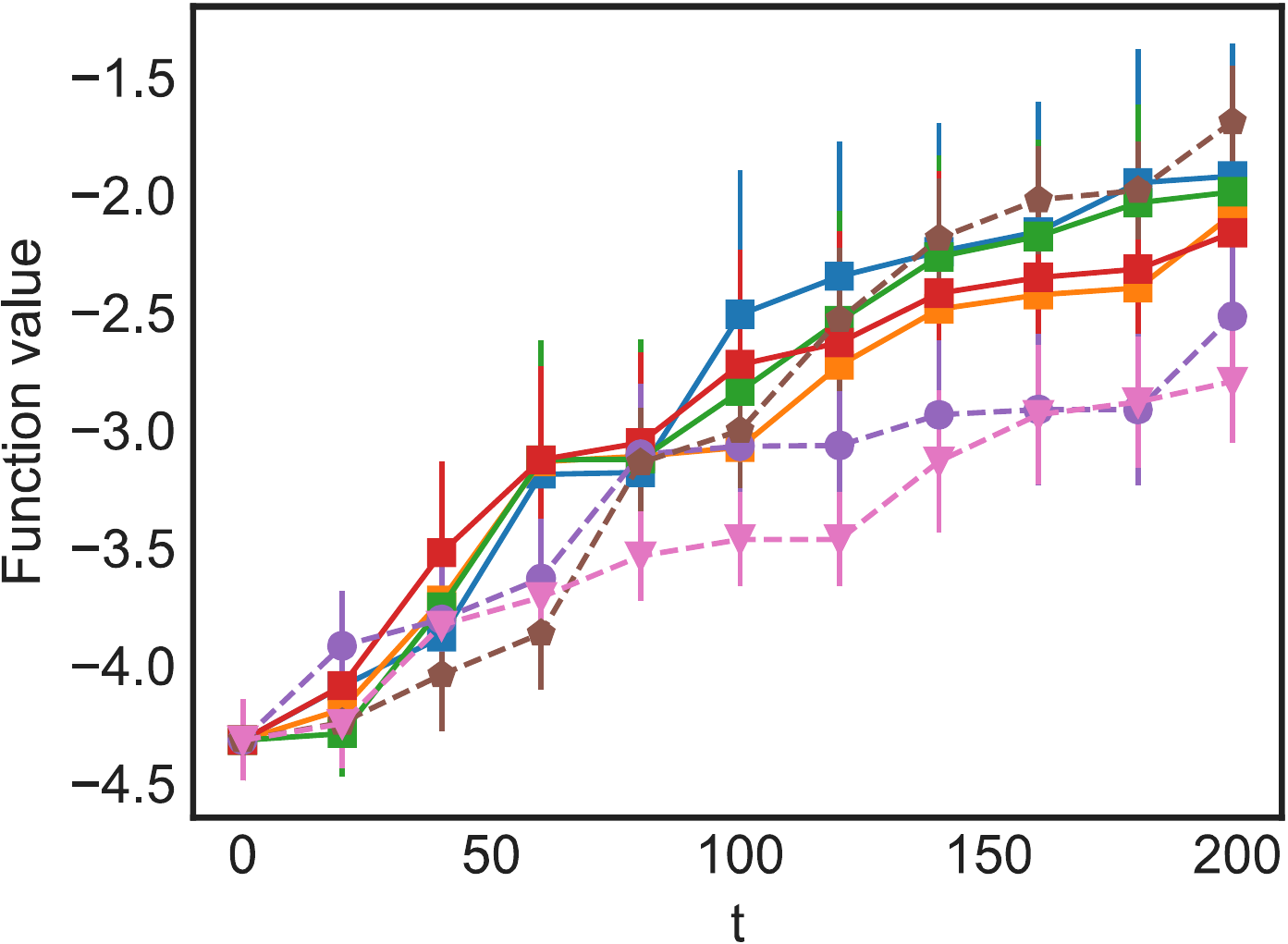}
    \caption{\textit{Ackley-5C}}
    \label{subfig:ackley5c16}
    \end{subfigure}
    
    \begin{subfigure}{0.33\linewidth}
     \centering
    \includegraphics[trim=0cm 0.cm 0cm  0.cm, clip, width=1.0\linewidth]{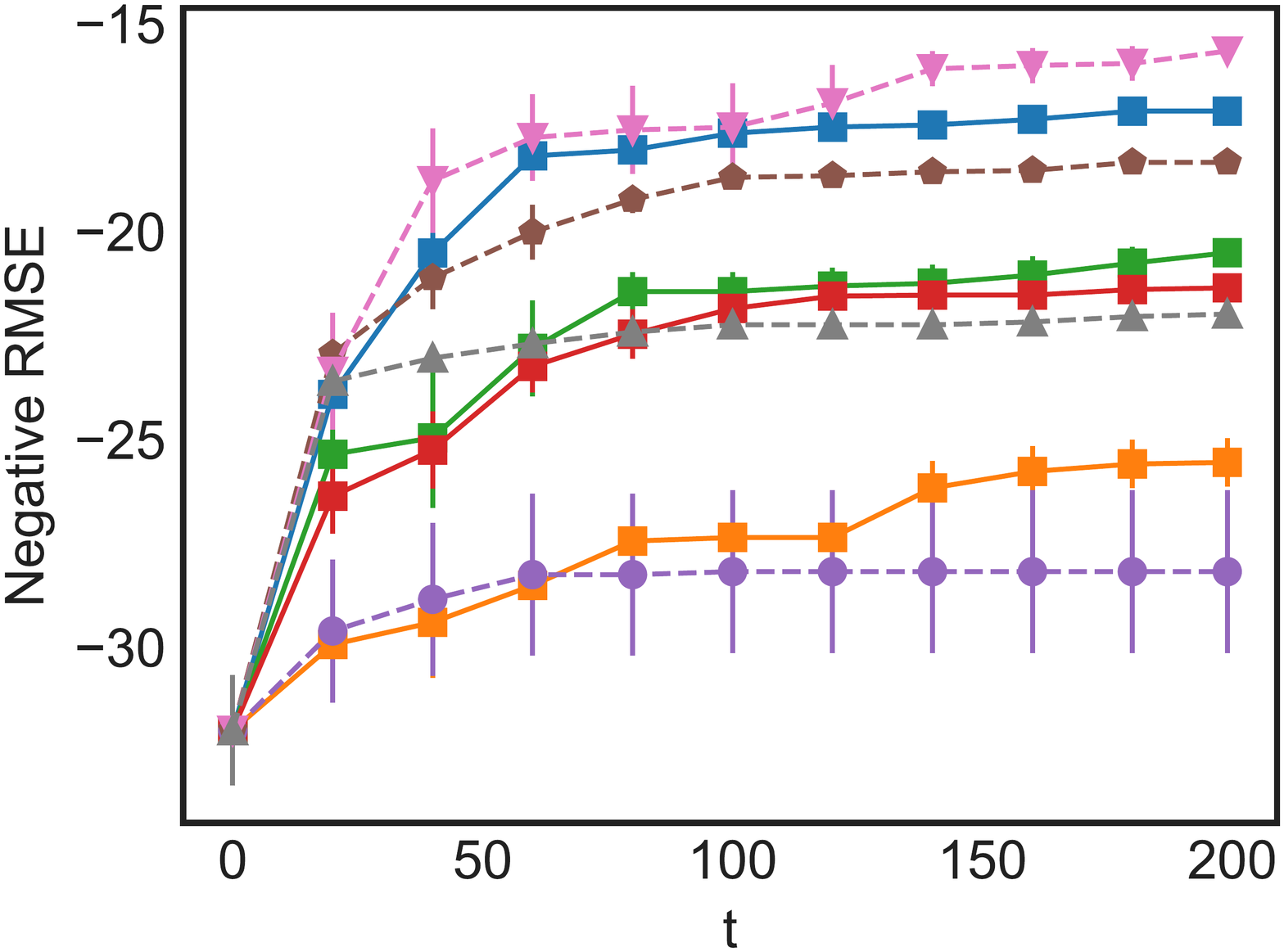}
    \caption{\textit{SVM-Boston}}
    \label{subfig:svmboston}
    \end{subfigure}
    \begin{subfigure}{0.33\linewidth}
     \centering
    \includegraphics[trim=0cm 0.cm 0cm  0.cm, clip, width=1.0\linewidth]{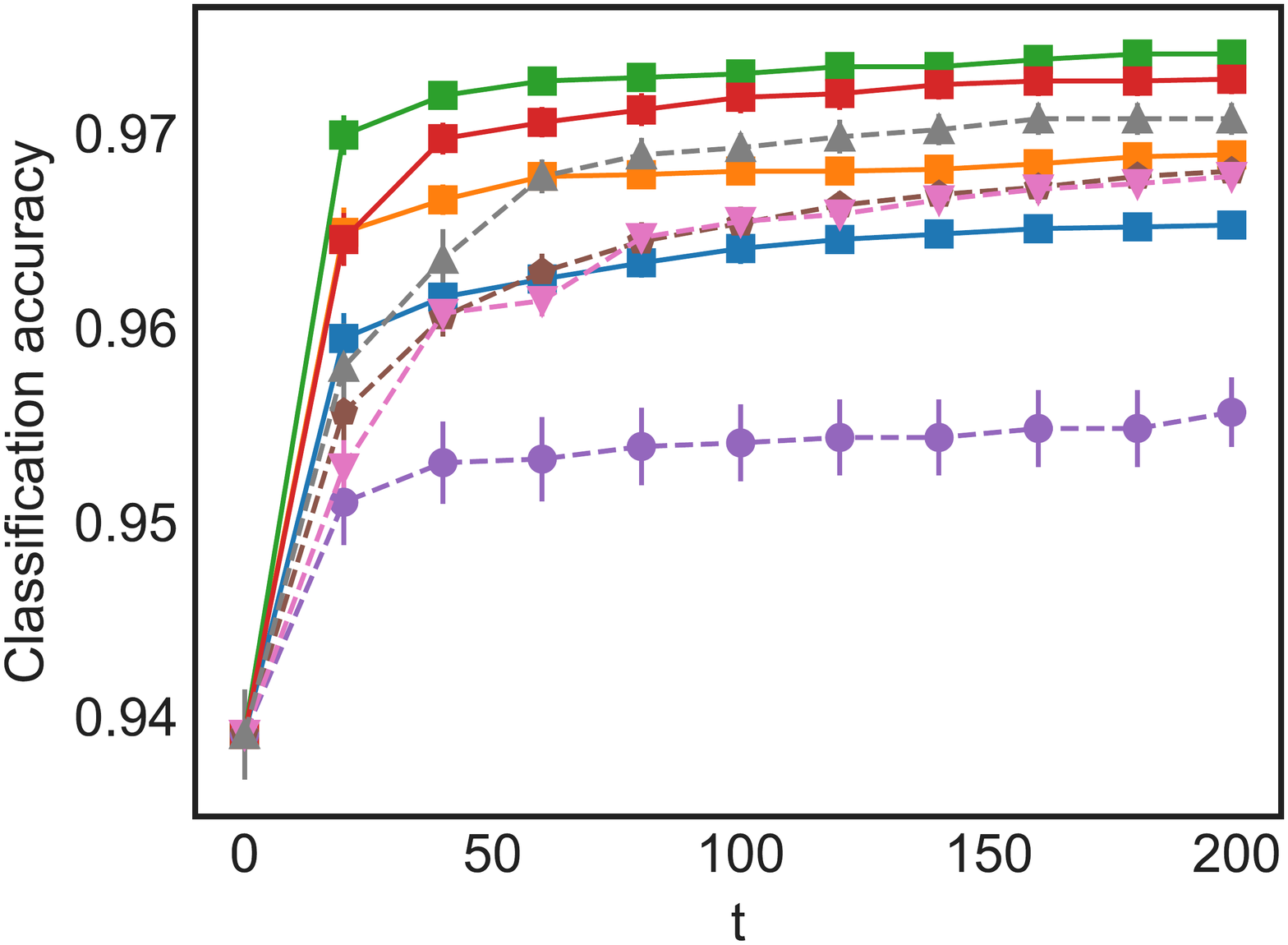}
    \caption{\textit{XG-MNIST}}
    \label{subfig:xgmnist}
    \end{subfigure}
    \begin{subfigure}{0.33\linewidth}
    \centering
    \includegraphics[trim=0cm 0.cm 0cm  0.cm, clip, width=1.0\linewidth]{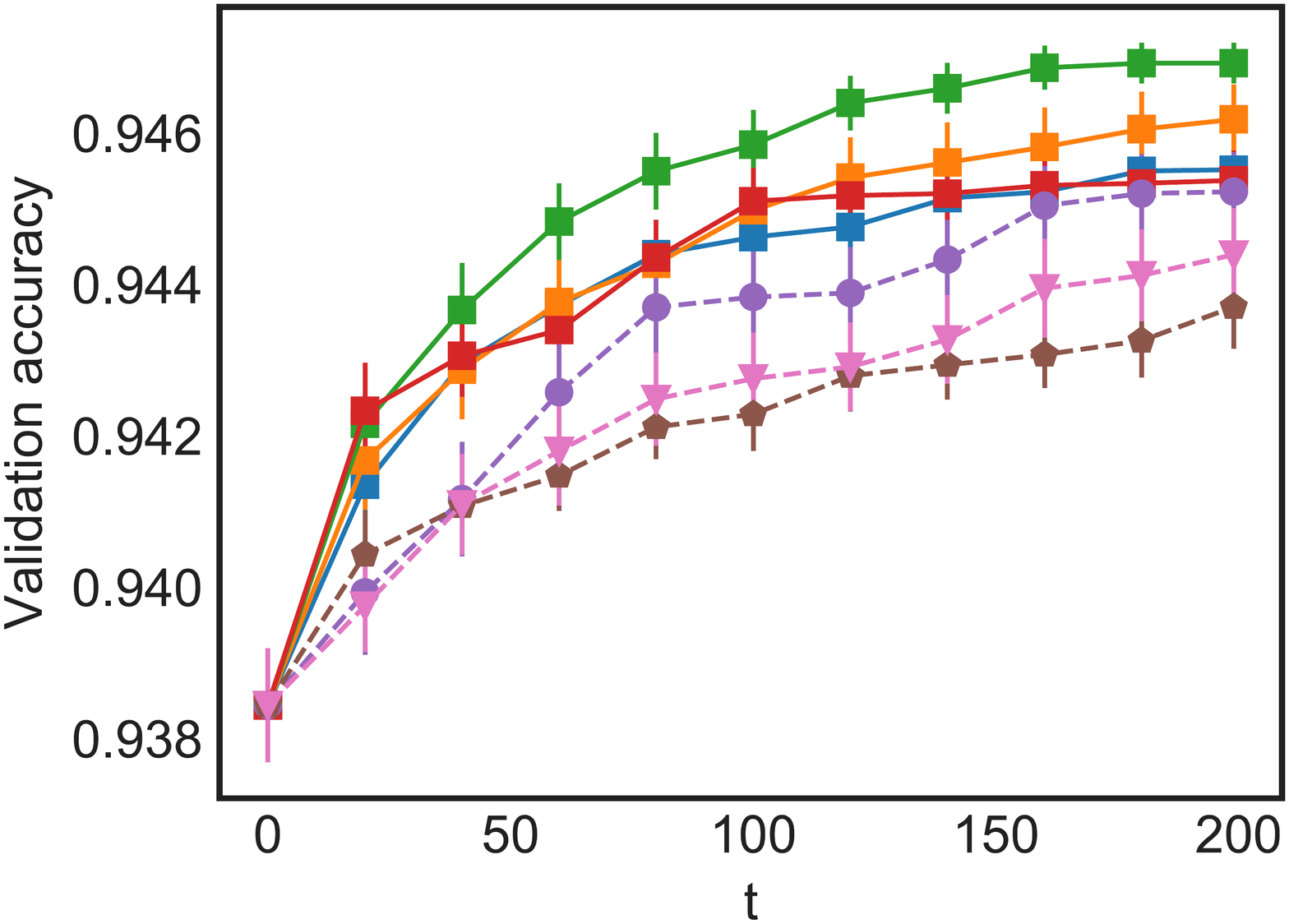}
    \caption{\textit{NAS-CIFAR10}}
    \label{subfig:nascifar10}
    \end{subfigure}
    
    \caption{Performance of CoCaBOs against existing methods on various tasks in the sequential setting ($b=1$).} \label{fig:sequential_sync_exps}
\end{figure*}

We compared CoCaBO against a range of existing methods which are able to handle problems with mixed type inputs: SMAC \cite{Hutter_2011Sequential}, TPE \cite{Bergstra_2011Algorithms}, GP-based Bayesian optimisation with one-hot encoding (One-hot BO) \cite{gpyopt2016} and EXP3BO \cite{gopakumar2018algorithmic_NIPS}. We did not compare to the concurrent category-specific approach of \cite{nguyen2019bayesian} because its code is not released yet. However, the method \cite{nguyen2019bayesian} is highly similar to EXP3BO and suffers the same limitations as EXP3BO. For all the baseline methods, we used their  publicly available Python packages\footnote{One-hot BO: \url{https://github.com/SheffieldML/GPyOpt}, SMAC: \url{https://github.com/automl/pysmac}, TPE: \url{https://github.com/hyperopt/hyperopt}, EXP3BO: \url{https://github.com/shivapratap/AlgorithmicAssurance_NIPS2018}}. 
CoCaBO and one-hot BO both use the UCB acquisition function \cite{Srinivas_2010Gaussian} with scale parameter $\kappa = 2.0$. In all experiments, we tested four different $\lambda$ values for our method\footnote{\url{https://github.com/rubinxin/CoCaBO_code}}: $\lambda=1.0, 0.5, 0.0, \text{auto}$, where $\lambda=\text{auto}$ means $\lambda$ is optimised as a hyperparameter. This leads to four variants of our method: CoCaBO-$1.0$, CoCaBO-$0.5$, CoCaBO-$0.0$ and CoCaBO-auto. We used a Mat\'ern, $\nu=\frac{5}{2}$, kernel for $k_x$, as well as for One-hot BO, and used the indicator-based kernel discussed in Section \ref{ssec:surrogate} for $k_h$.
For both our method and One-hot BO, we optimised the GP hyperparameters by maximising the log marginal likelihood every 10 iterations using multi-started gradient descent (see Appendix E for more details).

TPE is only used in the sequential setting ($b=1$) because its package HyperOpt does not provide a synchronous batch implementation. We started each optimisation method except EXP3BO with $24$ random initial points. EXP3BO requires much more initial data to start with because as mentioned in Section \ref{subsec:exp3bo}, it needs to fit a GP surrogate to each category. For example, if the problem involves 2 categorical variables, each of which has 3 categorical choices, EXP3BO needs to construct 9 independent GP surrogates by dividing the observation data into 9 subsets (one for each surrogate). When applying EXP3BO for our problems, we start it with 3 random initial points for each GP surrogate. 
For all the problems, the continuous inputs were normalised to $\mathbf{x} \in [-1, 1]^d$ and all experiments were conducted on a 36-core 2.3GHz Intel Xeon processor with 512 GB RAM.

\begin{figure*} [t]  
    \begin{subfigure}{0.33\linewidth}
     \centering
    \includegraphics[trim=0cm 0.cm 0cm  0.cm, clip, width=1.0\linewidth]{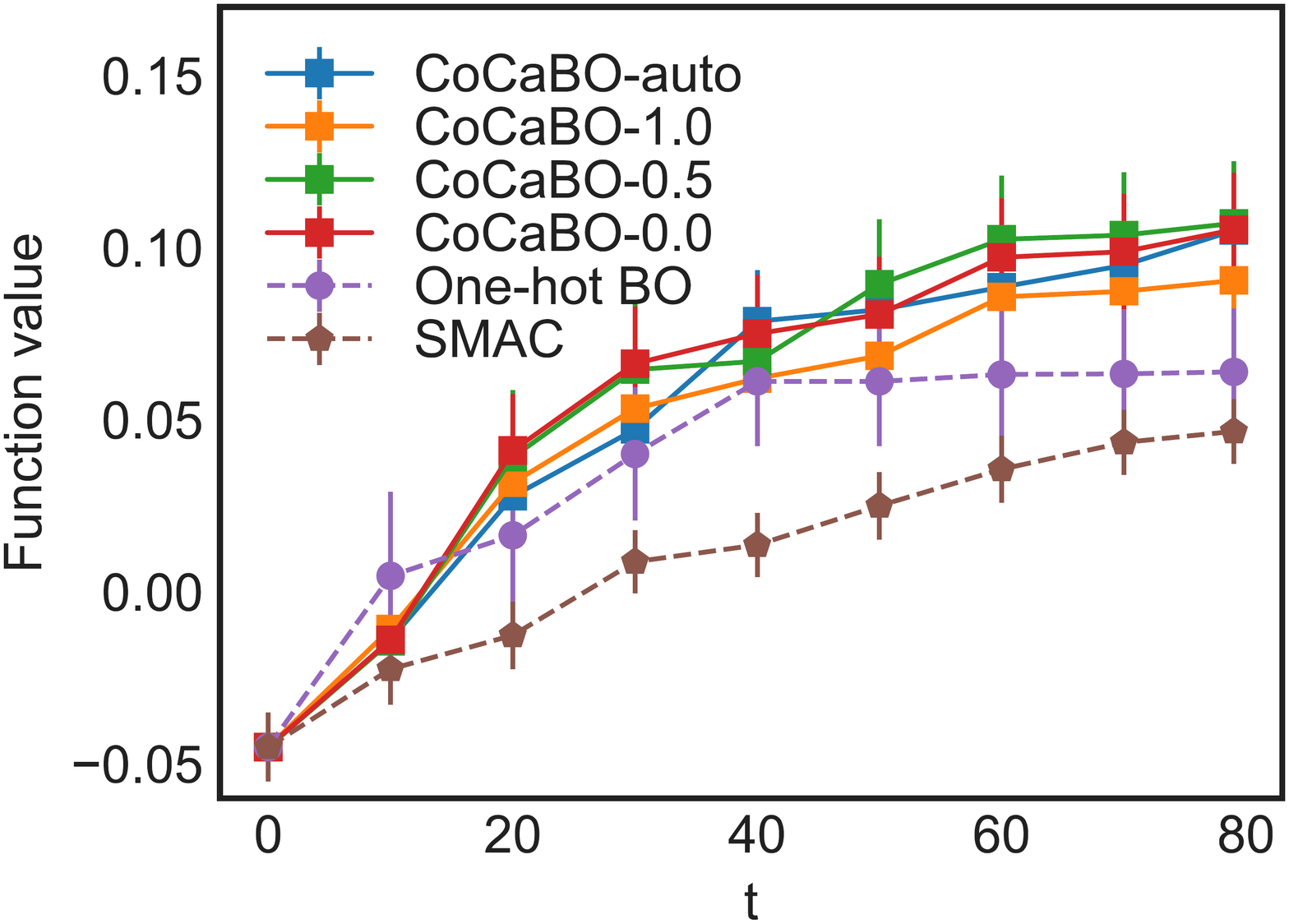}
    \caption{\textit{Func-2C}}
    \label{subfig:func2c4}
    \end{subfigure}
    \begin{subfigure}{0.33\linewidth}
    \centering
    \includegraphics[trim=0cm 0.cm 0cm  0.cm, clip, width=1.0\linewidth]{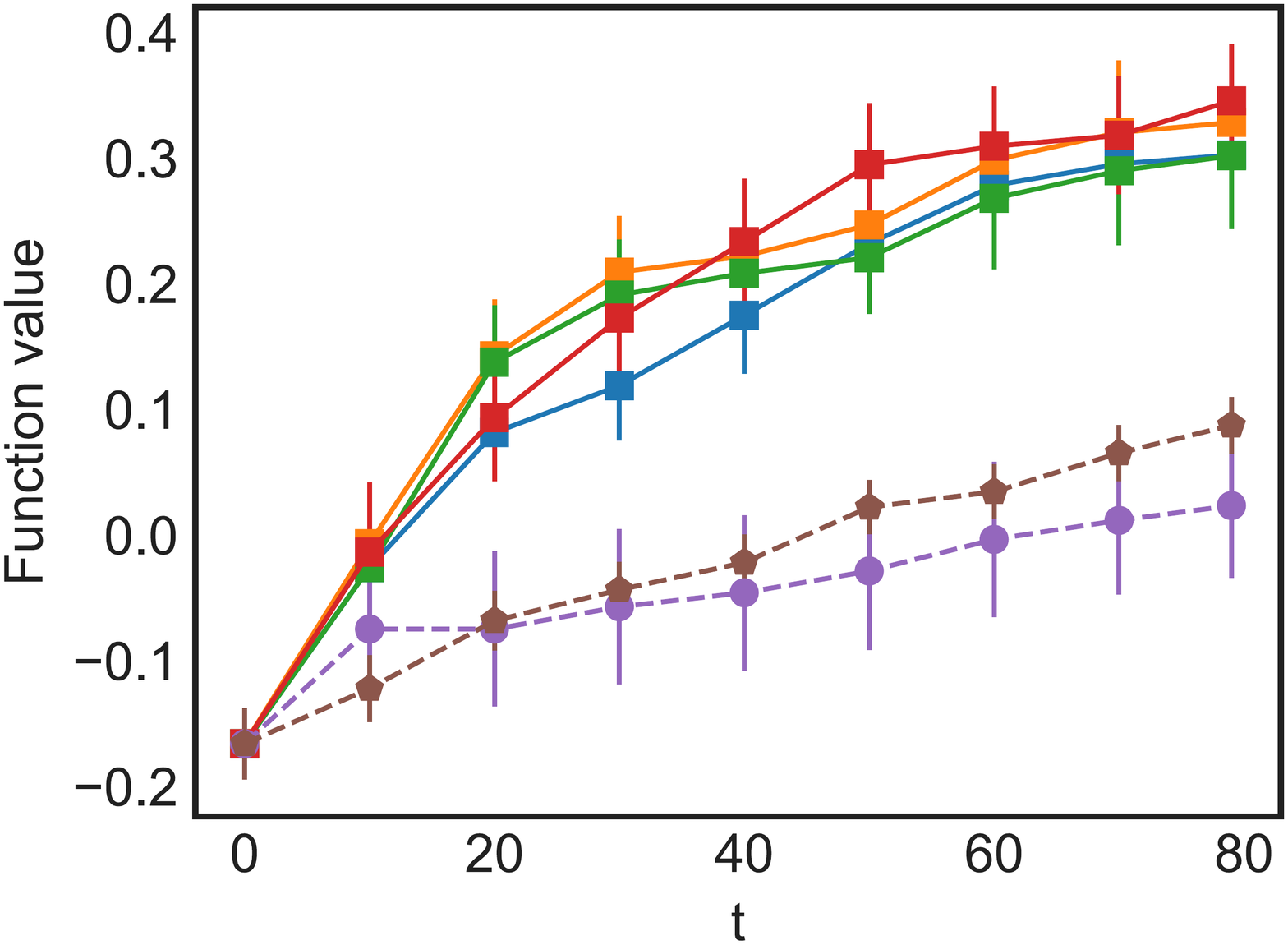}
    \caption{\textit{Func-3C}}
    \label{subfig:func3c4}
    \end{subfigure}
    \begin{subfigure}{0.33\linewidth}
     \centering
    \includegraphics[trim=0cm 0.cm 0cm  0.cm, clip, width=1.0\linewidth]{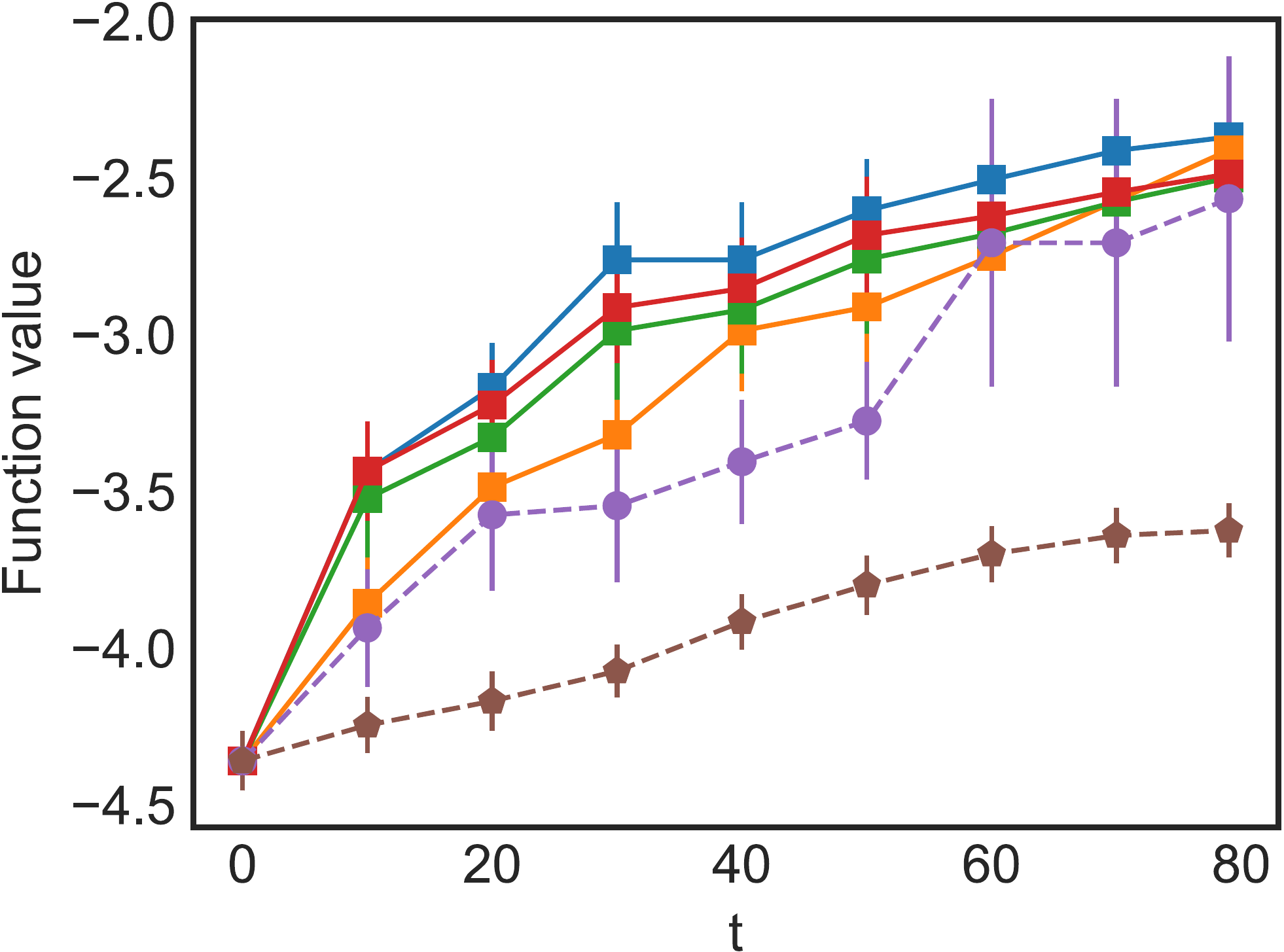}
    \caption{\textit{Ackley-5C}}
    \label{subfig:ackley5c164}
    \end{subfigure}
    
    \begin{subfigure}{0.33\linewidth}
     \centering
    \includegraphics[trim=0cm 0.cm 0cm  0.cm, clip, width=1.0\linewidth]{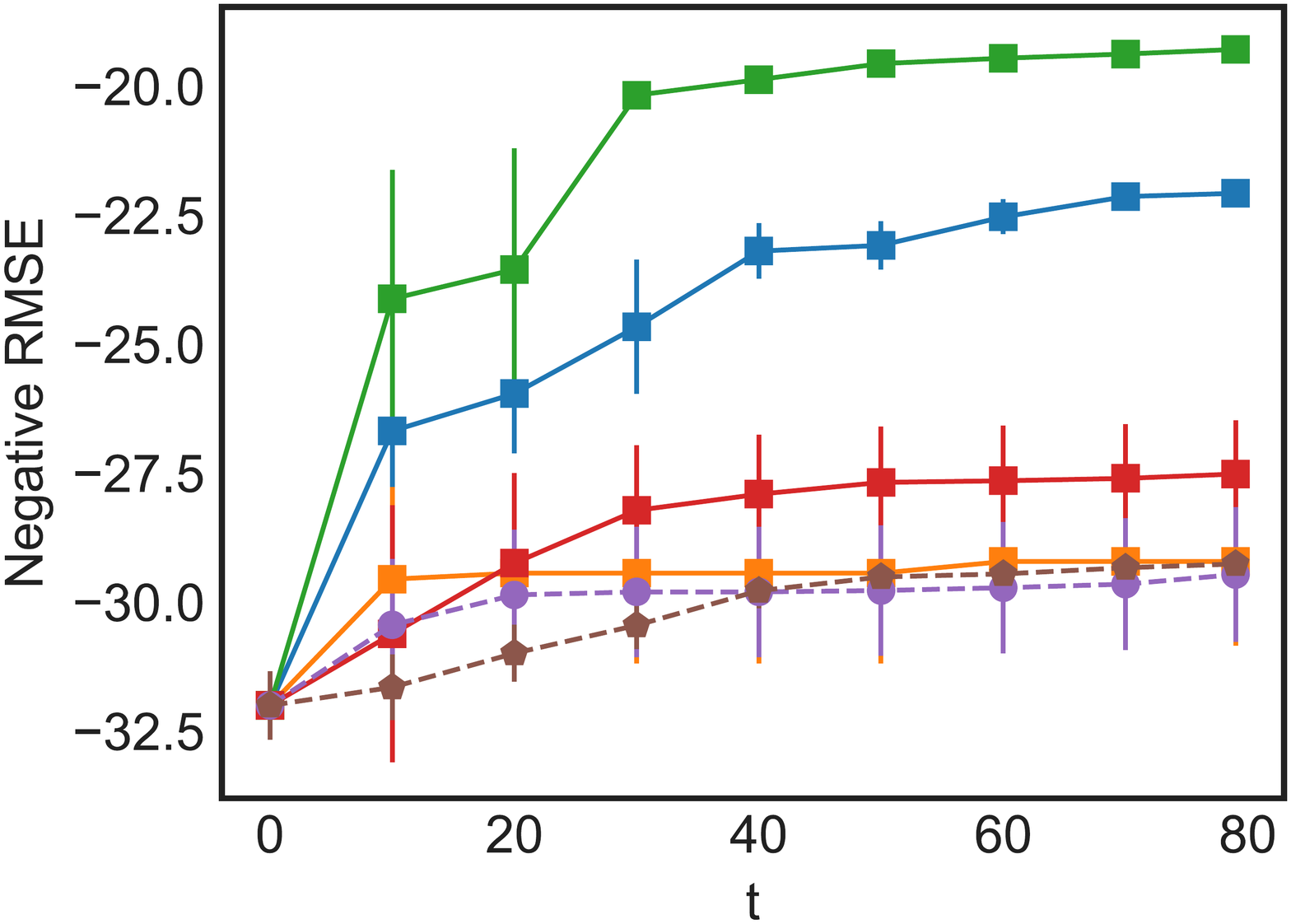}
    \caption{\textit{SVM-Boston}}
    \label{subfig:svmboston}
    \end{subfigure}
    \begin{subfigure}{0.33\linewidth}
     \centering
    \includegraphics[trim=0cm 0.cm 0cm  0.cm, clip, width=1.0\linewidth]{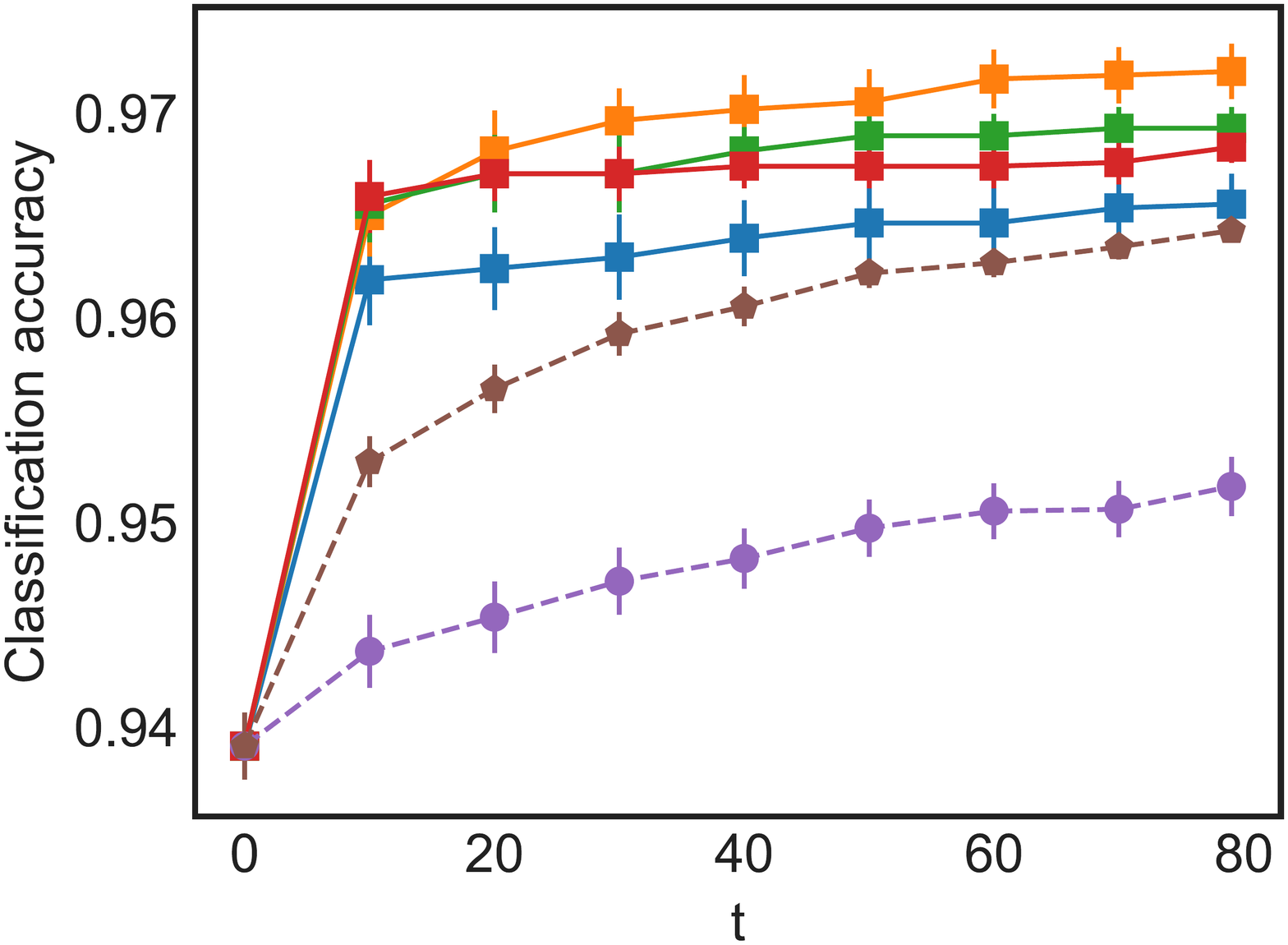}
    \caption{\textit{XG-MNIST}}
    \label{subfig:xgmnist}
    \end{subfigure}
    \begin{subfigure}{0.33\linewidth}
    \centering
    \includegraphics[trim=0cm 0.cm 0cm  0.cm, clip, width=1.0\linewidth]{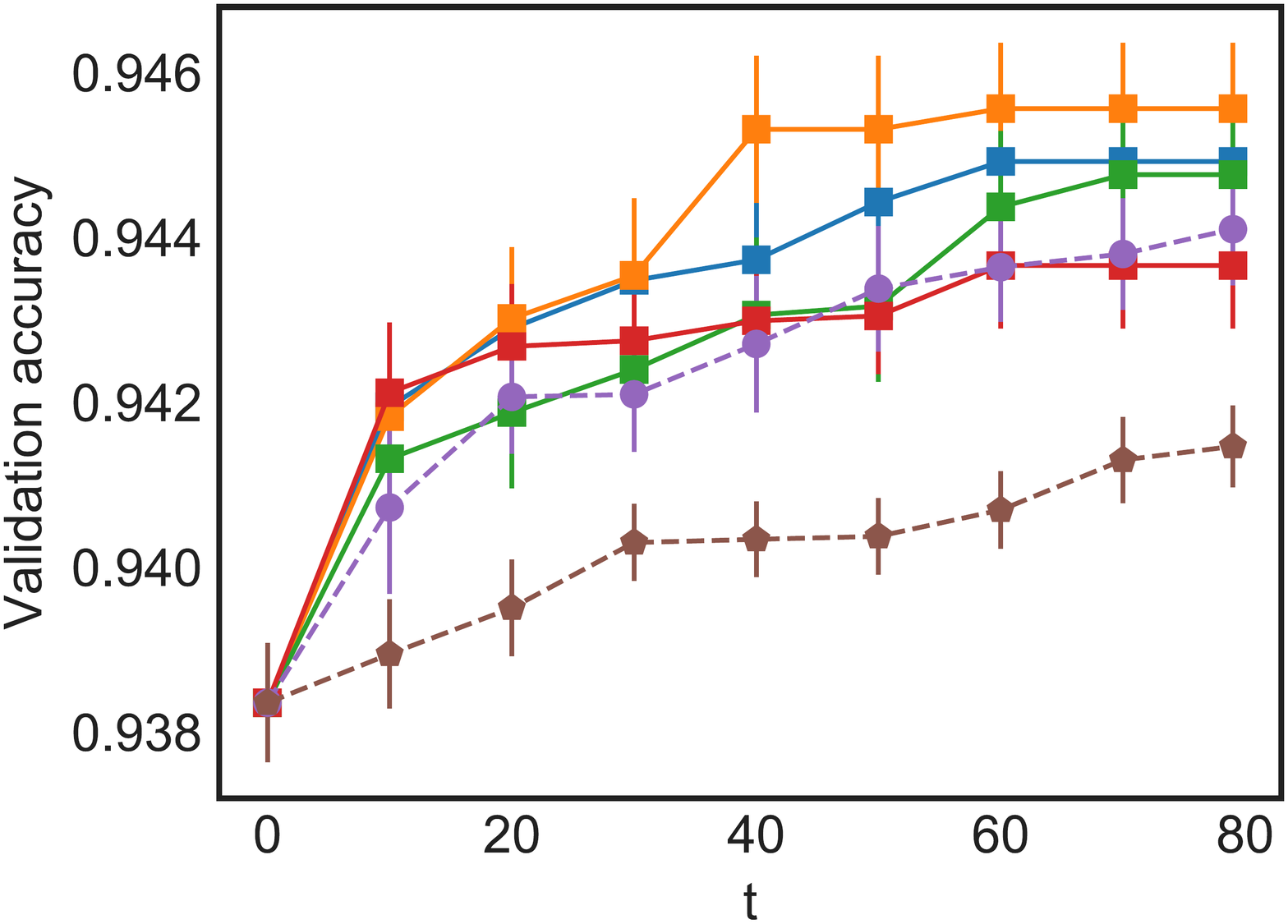}
    \caption{\textit{NAS-CIFAR10}}
    \label{subfig:nascifar10}
    \end{subfigure}
    
    \caption{Performance of CoCaBOs against existing methods on synthetic and real-world tasks in the batch setting ($b=4$).} \label{fig:batch_sync_exps}
\end{figure*}

We tested all these methods on a diverse set of synthetic and real problems:
\begin{itemize}
    \item \textit{Func-2C} is a test problem with $2$ continuous inputs ($d=2$) and $2$ categorical inputs ($c=2$). The categorical inputs control a linear combination of three $2$-dimensional global optimisation benchmark functions: beale, six-hump camel and rosenbrock. This is the function used for the illustration in Figure \ref{fig:CoCaBO_func2C_demo}; 
    \item \textit{Func-3C} is similar to \textit{Func-2C} but with $3$ categorical inputs which leads to more complicated combinations of the three functions;
    \item \textit{Ackley-$c$C}, with $c=\{2,3,4,5\}$ and $d=1$ is generated to test the performance of CoCaBO on problems with large numbers of categorical inputs and inputs with large numbers of categorical choices. Here, we convert $c$ dimensions of the $(c+1)$-dimensional Ackley function into $17$ categories each;
    \item  \textit{SVM-Boston} outputs the negative mean square test error of using a support vector machine (SVM) for regression on the Boston housing dataset \cite{Dua:2019}; 
    
    \item \textit{XG-MNIST} returns classification test accuracy of a XGBoost model \cite{chen2016xgboost} on MNIST \cite{lecun-mnisthandwrittendigit-2010};
    
    \item \textit{NAS-CIFAR10} performs the architecture search on convolutional neural network topology for CIFAR10 classification. We conducted the search using the NAS-Bench-101 dataset \cite{ying2019bench} and adopted the search space proposed in \cite{ying2019bench} which encodes each unique architecture with 5 categorical variables and 22 continuous variables. 
\end{itemize}

A brief summary of categorical and continuous inputs for the real-world problems is shown in Table 2 and a detailed description of all the problems is provided in Appendix G.

\subsection{Predictive performance of the CoCaBO posterior}
\label{ssec:surrogate-regression}

We first investigate the quality of the CoCaBO-auto surrogate by comparing its modelling performance against a standard GP with one-hot encoding. 
We train each model on $250$ uniformly randomly sampled data points and evaluate the predictive log likelihood on $100$ test data points. The mean and standard error over $20$ random initialisations are presented in Table \ref{tab:cocabo-vs-onehot-surrogate}. The results showcase the benefit of using the CoCaBO kernel over a kernel with one-hot encoded inputs, especially when the number of categorical inputs grows. The CoCaBO kernel, which allows it to learn a richer set of variations from the data, leads to consistently better out-of-sample predictions.

\subsection{Performance of CoCaBO in sequential setting}

We evaluated the optimisation performance of our proposed CoCaBO methods and other existing methods in the sequential setting. We  ran  each  sequential  optimisation method for $T= 200$ iterations. We performed $20$ random repetitions for synthetic problems and $10$ random repetitions for the real-world problems. The mean and standard error over all repetitions are presented in Figure \ref{fig:sequential_sync_exps}. For almost all synthetic and real-world problems, CoCaBO methods outperform other competing approaches with CoCaBO-$0.5$ and CoCaBO-auto consistently demonstrating the most competitive performance. 

Note that we did not apply EXP3BO for \textit{Ackley-5C} and \textit{NAS-CIFAR10} because both cases involve multiple categorical variables and each categorical variable has multiple categories, thus requiring EXP3BO to build a huge number of independent GP surrogates ($17^5$ and $3^5$ respectively) to model the objective problem. And this in turn demands a large amount of observation data to start the optimisation. Hence, EXP3BO is not suitable for problems involving \emph{multiple} categorical inputs with \emph{multiple} possible values. Even in the cases where EXP3BO is compared (Figure \ref{subfig:func2c}, \ref{subfig:func3c}, \ref{subfig:svmboston}, \ref{subfig:xgmnist}), CoCaBO outperformed EXP3BO. This justifies the gain in efficiency by using CoCaBO kernel to leverage all the data in a single surrogate in contrast with subdividing data by categories to learn multiple separate surrogates.

Another observation is that TPE performs relatively well in the case when the number of categorical combinations ( $\prod^{c}_{i}c N_i$) is small (e.g. \textit{Func-2C}(15), \textit{SVM-Boston}(16)) but performs clearly worse than CoCaBO when categorical combinations are large (e.g. \textit{Func-3C}($60$), \textit{NAS-CIFAR10}($243$), \textit{Ackley-$5$C}($17^5$)).

\subsection{Performance of CoCaBO in batch setting}\label{ssec:sequential_performance}

Now we move to experiments in the batch setting. We ran each batch optimisation with $b=4$ for $T=80$ iterations. The mean and standard error of the optimisation performance over $20$ random repetitions are presented in Figure \ref{fig:batch_sync_exps}. CoCaBO methods again show competitive performance over other batch methods with CoCaBO-$0.5$ and CoCaBO-auto again remaining the top performing $\lambda$ options. This supports our hypothesis that combining the sum and product kernels is beneficial for search over the mixed-type input space. Therefore, we recommend $\lambda=0.5$ or $\lambda=$auto as the default options for using our algorithm.

One point to note is that CoCaBO-$1.0$ (product kernel only) performs quite competitively in the batch setting for problems like XG-MNIST and NAS-CIFAR10 whose categorical variables have few category choices. At the same iteration number,the batch setting provides more sample data for the BO algorithm than the sequential setting, leading to more frequent overlapping categories in the data, especially when the category choices are small. Hence, the product kernel, which captures the couplings between the continuous and categorical data, can perform more effectively.  




\section{Conclusion}

\label{sec:conclusion}

Existing BO literature uses one-hot transformations or hierarchical approaches to encode real-world problems involving mixed continuous and categorical inputs. We presented a solution from a novel perspective, called Continuous and Categorical Bayesian Optimisation (CoCaBO), that harnesses the strengths of multi-armed bandits and GP-based BO to tackle this problem. Our method uses a new kernel structure, which allows us to capture information within categories as well as across different categories. This leads to more efficient use of the acquired data and improved modelling power. We extended CoCaBO to the batch setting, enabling parallel evaluations at each stage of the optimisation. CoCaBO demonstrated strong performance over existing methods on a variety of synthetic and real-world optimisation tasks with \emph{multiple} continuous and categorical inputs. We find CoCaBO to offer a very competitive alternative to existing approaches.

\bibliographystyle{icml}
\bibliography{mabbobib}

\end{document}



\icmltitlerunning{Bayesian Optimisation over Multiple Continuous and Categorical Inputs}

\twocolumn[

\icmltitle{Supplement to Bayesian Optimisation over Multiple Continuous and Categorical Inputs}

\icmlsetsymbol{equal}{*}

\begin{icmlauthorlist}

\icmlauthor{Binxin Ru$\text{}^*$} {ox} 
\icmlauthor{Ahsan S. Alvi$\text{}^*$} {ox}
\icmlauthor{Vu Nguyen} {ox}
\icmlauthor{Michael A. Osborne } {ox}
\icmlauthor{Stephen J Roberts } {ox}

\end{icmlauthorlist}

\icmlaffiliation{ox}{University of Oxford}

\icmlcorrespondingauthor{Binxin Ru}{robin@robots.ox.ac.uk}

\icmlkeywords{Machine Learning, ICML}

\vskip 0.3in
]

\printAffiliationsAndNotice{\icmlEqualContribution} 


\appendix 

\section{Notation summary}
 Please refer to Table \ref{tab:Notation-list} for a description of the notations used in our paper.
\begin{table*}[t]
  \caption{Notation list} 
  \label{tab:Notation-list}
  \centering
    \begin{tabular}{lll}
    \toprule
    \textbf{Notation}    & \textbf{Type}     & \textbf{Meaning} \\\midrule
    $\sigma_{l}^{2},\sigma^{2}$ & scalar & \makecell[l]{lengthscale for RBF kernel, \\noise output variance (or measurement noise)}\\\midrule
 
    
    $\mathcal{X}\in\mathbb{R}^{d}$ & search domain & continuous search space where $d$
is the dimension\\\midrule

    $d$ & scalar & dimension of the continuous variable\\\midrule
    
    $c$ & scalar & dimension of categorical variables\\\midrule
    
    $\mathbf{x}_{t}$ & vector & a continuous selection by BO at iteration $t$\\\midrule
    
    $N_{c}$ & scalar & number of choices for categorical variable $c$\\\midrule
    
    $\mathbf{h}_{t}=[h_{t,1},...,h_{t,c}]$ & vector & vector of categorical variables\\\midrule
 
 
    $\mathbf{z}_{t}=[\mathbf{x}_{t},\mathbf{h}_{t}]$ & vector & \makecell[l]{hyperparameter input including continuous and \\ categorical variables}\\\midrule
 
    $\mathcal{D}_{t}$ & set & observation set $D_{t}=\{z_{i},y_{i}\}_{i=1}^{t}$\\\midrule
 
    \end{tabular}
\end{table*}


\section{EXP3}
\label{app:exp3_algorithm}
EXP3 is a method proposed in \cite{auer2002nonstochastic} to deal with the non-stochastic, adversarial multi-arm bandit problem which is a more general setting and can model any form of non-stationarity \cite{allesiardo2017non}. In such problem setting, the rewards $g_t$ are chosen by an adversary at each iteration.
For a categorical variable $h$ with $N$ categories and bounded reward ($g_t \leq 1$), the regret bound for EXP3 algorithm is 
\begin{align}
    \mathcal{R}^{EXP3}_T &= \max_{h^* \in \{1, \ldots, N \}} \sum_{t}^T g_t(h^*) - \mathbb{E} \left[ \sum_t^T g_t(h) \right] \nonumber \\
    &\leq 2.63 \sqrt{T N \log (N)}
\end{align}
where $h^*$ is the best single action over all rounds and $g_t(\cdot)$ is the reward function. Please refer to \cite{auer2002nonstochastic} for detailed derivation.


\section{Regret bound for EXP3 in CoCaBO}

Assume we have $c$ categorical variables $\mathbf{h}=[h_1, \ldots, h_c]$ and each categorical variables $h_j$, determined by an agent, can take one of $N$ categories. A simplified version of the cumulative regret of such multi-agent MAB setting after $T$ iterations can be written as:
\begin{align}
    \sum_{j}^c \mathcal{R}^j_T = &
    \sum_{j}^c \sup_{\mathbf{h}_{\setminus j}} \Bigg\{ \max_{h_j^* \in \{1, \ldots, N \}} \sum_{t}^T g_t(h_j^*, \mathbf{h}_{\setminus j}) \\
    &- \mathbb{E} \left[ \sum_t^T g_t(h_j, \mathbf{h}_{\setminus j}) \right] \Bigg\}
\end{align}

where $\mathcal{R}^j_T$ is the cumulative regret of agent $j$ when we consider the decisions by the other agents $\mathbf{h}_{\setminus j}$ are controlled by the adversary. $g_t(\cdot)$ is the reward achieved by the joint decisions of all the agents and is equal to the normalised (between $[0,1]$) objective function value in our case.

We first look at the cumulative regret for a single agent when when the decision of all the other agents (i.e. the value of all the other categorical variables) are fixed to $\mathbf{h}_{\setminus j}$:
\begin{align}\label{eq:cum_regret_one_agent}
    \mathcal{R}^j_T = & \sup_{\mathbf{h}_{\setminus j}} \Bigg\{ \max_{h_j^* \in \{1, \ldots, N \}} \sum_{t}^T g_t(h_j^*, \mathbf{h}_{\setminus j}) \\
    &- \mathbb{E} \left[ \sum_t^T g_t(h_j, \mathbf{h}_{\setminus j}) \right] \Bigg\}
\end{align}

At iteration $t$ with given continuous inputs, the adversary's reward function $g_t(h_j, \mathbf{h}_{\setminus j})$ (with fixed $\mathbf{h}_{\setminus j}$) is equivalent to $\hat{g}_t(h_j)$ where $g_t(\cdot)$ has $N^c$ arbitrary outputs while $\hat{g}_t(\cdot)$ has only $N$ arbitrary outputs \cite{carlucci2020manas}. By substituting $\hat{g}_t(h_j) =  g_t(h_j,\mathbf{h}_{\setminus j})$ in Equation \eqref{eq:cum_regret_one_agent}, we reduce the setting for a single agent to the EXP3 setting \cite{auer2002nonstochastic}:

\begin{align}\label{eq:cum_regret_one_agent}
    \mathcal{R}^j_T & = 
    \sup_{\mathbf{h}_{\setminus j}} \Bigg\{ \max_{h_j^* \in \{1, \ldots, N \}} \sum_{t}^T \hat{g}_t(h^*_j) \nonumber
    - \mathbb{E} \left[ \sum_t^T \hat{g}_t(h_j) \right] \Bigg\} \nonumber \\
    & = \sup_{\mathbf{h}_{\setminus j}} \Bigg\{ \mathcal{R}^{EXP3, j}_T \Bigg\} \leq \sup_{\mathbf{h}_{\setminus j}} \Bigg\{ 2.63 \sqrt{T N \log (N)} \Bigg\}
\end{align}

Therefore, the cumulative regret of our multi-agent EXP3 setting after $T$ iterations has the bound:
\begin{align}
    \sum_{j}^c \mathcal{R}^j_T \leq 2.63 c \sqrt{T N \log (N)}
\end{align}

which is sub-linear as $\lim_{T \rightarrow \infty} \frac{\sum_{j}^c \mathcal{R}^j_T}{T} = 0$ and increases with the number of categorical variables $c$ and the number of categories for each variable.

\section{Categorical kernel relation with RBF}
\label{app:cat-kernel}
In this section we discuss the relationship between the categorical kernel we have proposed and a RBF kernel. 
Our categorical kernel is reproduced here for ease of access:
\begin{equation}
\label{eq:cat-kernel2}
    k_h(\mathbf{h}, \mathbf{h}') =  \frac{\sigma^2}{c} \sum_{i=1}^{c}\mathbb{I}(h_i - h_i').
\end{equation}
Apart from the intuitive argument, that this kernel allows us to model the degree of similarity between two categorical selections, this kernel can also be derived as a special case of an RBF kernel.
Consider the standard RBF kernel with unit variance evaluated between two scalar locations $a$ and $a'$:
\begin{equation}
\label{eq:basic-rbf}
    k(a, a') = \exp\left(-\frac{1}{2} \frac{(a - a')^2}{l^2}   \right).
\end{equation}
The lengthscale in Eq, \ref{eq:basic-rbf} allows us to define the similarity between the two inputs, and, as the lengthscale becomes smaller, the distance between locations that would be considered similar (i.e. high covariance) shrinks. 
The limiting case $l\rightarrow 0$ states that if two inputs are not exactly the same as each other, then they provide no information for inferring the GP posterior's value at each other's locations.
This causes the kernel to turn into an indicator function as in Eq. \ref{eq:cat-kernel2} above \cite{kulis2011revisiting}:
\begin{equation}
     k(a, a') = 
\begin{cases}
    1,& \text{if } a = a'\\
    0,              & \text{otherwise}.
\end{cases}
\end{equation}
By adding one such RBF kernel with $l\rightarrow 0$ for each categorical variable in $h$ and normalising the output we arrive at the form in Eq. \ref{eq:cat-kernel2}.

\section{Learning the hyperparameters in the CoCaBO kernel}
\label{app:learning}
We present the derivative for estimating the variable $\lambda$ in our CoCaBO kernel.
\begin{align}
\label{eq:app-CoCaBO-kernel}
    k_z(\mathbf{z}, \mathbf{z}') = &   (1-\lambda)\left ( k_h(\mathbf{h}, \mathbf{h}') + k_x(\mathbf{x}, \mathbf{x}') \right) \nonumber\\
    &+ \lambda  k_h(\mathbf{h}, \mathbf{h}')k_x(\mathbf{x}, \mathbf{x}').
\end{align}
The hyperparameters of the kernel are optimised by maximising the log marginal likelihood (LML) of the GP 
\begin{equation}
\label{eq:app-LML-1}
    \theta^* = \arg \max_\theta \mathcal{L}(\theta, \mathcal{D}),
\end{equation}
where we collected the the hyperparameters of both kernels as well as the CoCaBO hyperparameter into $\theta = \{\theta_h, \theta_x, \lambda\}$. 
The LML and its derivative are defined as
\begin{equation}
    \label{eq:app:LML}
    \mathcal{L}(\theta) = -\frac{1}{2}\mathbf{y}^\intercal \mathbf{K}^{-1}\mathbf{y} 
    - \frac{1}{2}\log |\mathbf{K}| + \text{constant}
\end{equation}
\begin{equation}
    \label{eq:app:LMLderiv}
    \frac{\partial\mathcal{L}}{\partial \theta} = \frac{1}{2}\left(\mathbf{y}^\intercal \mathbf{K}^{-1}\frac{\partial\mathbf{K}}{\partial\theta}\mathbf{K}^{-1}\mathbf{y} 
    -\text{tr}\left( \mathbf{K}^{-1} \frac{\partial\mathbf{K}}{\partial\theta} \right)\right),
\end{equation}
where $\mathbf{y}$ are the function values at sample locations and $\mathbf{K}$ is the kernel matrix of $k_z(\mathbf{z}, \mathbf{z}')$ evaluated on the training data  \cite{Rasmussen_2006gaussian}.

Optimisation of the LML was performed via multi-started gradient descent. 
The gradient in Equation \ref{eq:app:LMLderiv} relies on the gradient of the kernel $k_z$ w.r.t. each of its parameters:
\begin{align}
    \frac{\partial k_z}{\partial\theta_h} &=
    (1-\lambda)\frac{\partial k_h}{\partial\theta_h} + \lambda k_x \frac{\partial k_h}{\partial\theta_h}   \\ 
    \frac{\partial k_z}{\partial\theta_x} &=
    (1-\lambda)\frac{\partial k_x}{\partial\theta_x} + \lambda \frac{\partial k_x}{\partial\theta_x} k_h  \\
    \frac{\partial k_z}{\partial\lambda} &=
    -(k_h + k_x ) +  k_h k_x ,
\end{align}
where we used the shorthand $k_z =k_z(\mathbf{z}, \mathbf{z}')$, $k_h = k_h(\mathbf{h}, \mathbf{h}')$ and $k_x = k_x(\mathbf{x}, \mathbf{x}')$.

 
\section{Kriging Believer}
\label{app:kb_explain}

The Kriging Believer (KB) is a method proposed in \cite{Ginsbourger_2010Kriging} to sequentially select batch points in the continuous space. In KB, as the name suggested, we fully trust the predictive posterior and use the posterior mean $\mu(\mathbf{x}_t^{(1)})$ at a selected batch location $\mathbf{x}_t^{(1)}$ as a proxy for the true function value $f(\mathbf{x}_t^{(1)})$. We then augment the observation data $D_{t-1}$ with this hallucinated data $\{ \mathbf{x}_t^{(1)}, \mu(\mathbf{x}_t^{(1)}) \}$ to update the surrogate model as well as the the acquisition function. The next point in the batch is then selected by maximising the updated acquisition function. This process repeats until all $b$ points in the batch are selected as shown in Algorithm \ref{alg:KB}.

\begin{algorithm}[H]
	    \caption{Kriging Believer}\label{alg:KB}
	\begin{algorithmic}[1]
		\vspace{0.5em}
		\STATE {\bfseries Input:}  Observation data $\mathcal{D}_{t-1}$, batch size $b$
		\STATE {\bfseries Output:} The batch points $\mathcal{B}_t = \{\mathbf{x}_t^{(1)}, \ldots, \mathbf{x}_t^{(b)}\}$
		\STATE $\mathcal{D}'_{t-1}=\mathcal{D}_{t-1}$
        \FOR{$j=1, \dots, b$}
            \STATE $\mathbf{x}_t^{(j)}=\argmax \alpha(\mathbf{x}\vert \mathcal{D}^{'}_{t-1})$
            \STATE Compute $\mu(\mathbf{x}_t^{(j)})$ 
        	\STATE $\mathcal{D}'_{t-1} \leftarrow \mathcal{D}'_{t-1} \cup (\mathbf{x}_t , \mu(\mathbf{x}_t^{(j)}))$
    	\ENDFOR
	\end{algorithmic}
\end{algorithm}

\section{Description of the optimisation problems}
\label{app:test_problems}

\subsection{Synthetic test functions}
\label{app:syn_func_description}

We generated several synthetic test functions: \textit{Func-2C}, \textit{Func-3C} and a \textit{Ackley-$c$C} series, whose input spaces comprise both continuous variables and multiple categorical variables. Each of the categorical inputs in all three test functions have multiple values. 
\paragraph{\textit{Func-2C}} is a test problem with $2$ continuous inputs ($d=2$) and $2$ categorical inputs ($c=2$). The categorical inputs decide the linear combinations between three $2$-dimensional global optimisation benchmark functions: beale (bea), six-hump camel (cam) and rosenbrock (ros)\footnote{The analytic forms of these functions are available at \url{https://www.sfu.ca/~ssurjano/optimization.html}}.

\paragraph{\textit{Func-3C}} is similar to \textit{Func-2C} but with $3$ categorical inputs ($c=3$) which leads to more complicated linear combinations among the three functions. 

\paragraph{\textit{Ackley-$c$C}} comprises $c=\{2,3,4,5\}$ categorical inputs and $1$ continous input ($d=1$). Here, we convert $c$ dimensions of the $c+1$-dimensional Ackley function into $17$ categories each.

The value range for both continuous and categorical inputs of these functions are summarised in Table \ref{table:sync_problem_input_details}.

\begin{table*}
  \caption{Continuous and categorical input range of the synthetic test functions}
  \vspace{8pt}
  \label{table:sync_problem_input_details}
    \centering
    \begin{tabular}{lll}
    \toprule
    Function $f$ &  Inputs $\mathbf{z}=[\mathbf{h},\mathbf{x}]$  & Input values  \\ \midrule
    \multirow{3}{*}{\thead{\textit{Func-2C} \\($d=2$, $c=2$)}} 
    & $h_1$     & $\{ \text{ros}(\mathbf{x}), \text{cam}(\mathbf{x}), \text{bea}(\mathbf{x}) \}$      \\ 
    & $h_2$     & $\{ + \text{ros}(\mathbf{x}), +\text{cam}(\mathbf{x}), +\text{bea}(\mathbf{x}), +\text{bea}(\mathbf{x}), +\text{bea}(\mathbf{x}) \}$ \\ 
    & $\mathbf{x}$    & $[-1,1]^2$    \\
    \midrule
    \multirow{4}{*}{\thead{\textit{Func-3C}\\ ($d=2$, $c=3$)}} 
    & $h_1$     & $\{ \text{ros}(\mathbf{x}), \text{cam}(\mathbf{x}), \text{bea}(\mathbf{x}) \}$      \\ 
    & $h_2$     & $\{ + \text{ros}(\mathbf{x}), +\text{cam}(\mathbf{x}), +\text{bea}(\mathbf{x}), +\text{bea}(\mathbf{x}), +\text{bea}(\mathbf{x}) \}$ \\ 
    & $h_3$     & $\{ + 5 \times \text{cam}(\mathbf{x}), + 2 \times \text{ros}(\mathbf{x}), + 2 \times \text{bea}(\mathbf{x}), + 3 \times \text{bea}(\mathbf{x}) \}$ \\
    & $\mathbf{x}$    & $[-1,1]^2$    \\\midrule
    \multirow{3}{*}{\makecell[c]{\textit{Ackley-$c$C} for \\ $c=\{2, 3, 4, 5\}$\\ ($d=1$, $N_i=17$)}} 
    &  $h_i$ \hspace{1pt} for    & $\{z_i = -1 + 0.125 \times (j-1), \text{\hspace{1pt} for \hspace{1pt}} j=1,2, \dots, 17 \}$    \\ 
    & $i=1,2, \dots, 5$     &  \\
    & $\mathbf{x}$    & $[-1,1]$    \\ 
    \bottomrule
\end{tabular}
\end{table*}

\subsection{Real-world problems}
\label{app:real_probelm_description}

\begin{table*}
  \caption{Continuous and categorical input ranges of the real-world problems}
  \vspace{8pt}
  \label{table:real_problem_inputs_details}
    \centering
    \begin{tabular}{lll}
    \toprule
    Problems &  Inputs $\mathbf{z}=[\mathbf{h},\mathbf{x}]$  & Input values  \\ \midrule
    \multirow{3}{*}{\thead{\textit{SVM-Boston} \\($d=3$, $c=3$)}} 
    & kernel type $h_1$     & $\{$linear, poly, RBF, sigmoid$\}$      \\ 
    & kernel coefficient $h_2$  & $\{$scale, auto $\}$ \\ 
    & shrinking $h_3$   & $\{$shrinking on, shrinking off$\}$ \\
    & penalty parameter $x_1$    & $[0,10]$    \\
    & tolerance for stopping $x_2$    & $10^{[10^{-6},1]}$    \\
    & \thead{lower bound of the fraction \\ of support vector $x_3$}    & $[0,1]$    \\
    \midrule
    \multirow{4}{*}{\thead{\textit{XG-MNIST} \\($d=5$, $c=3$)}} 
    & booster type $h_1$     & $\{$gbtree, dart$\}$      \\ 
    & grow policies $h_2$  & $\{$depthwise, loss$\}$ \\ 
    & training objective $h_3$   & $\{$softmax, softprob$\}$\\
    & learning rate $x_1$    & $[0,1]$    \\
    & maximum dept $x_2$    & $[1,2,\dots,10]$    \\
    & minimum split loss $x_3$    & $[0,10]$    \\
    & subsample $x_4$    & $[0.001,1]$    \\
    & regularisation $x_5$    & $[0,5]$    \\
    \midrule
    \multirow{4}{*}{\thead{\textit{NAS-CIFAR10}\\ ($d=22$, $c=5$)}} 
    & operations for the 5 intermediate  &       \\ 
    & nodes in the DAG $h_1, \ldots, h_5$ & $\{$3x3 conv, 1x1 conv, and 3x3 max-pool$\}$ \\ 
    & probability values for the 21 possible  &     \\
    & edges in the DAG $x_1, \ldots, x_{21}$  & $[0,1]$    \\
    & Number of edges present in the DAG $x_{22}$ & $[0, 9]$    \\  
    \midrule
    \multirow{4}{*}{\thead{\textit{NN-Yacht}\\ ($d=3$, $c=3$)}} 
    & activation type $h_1$     & $\{$ReLU, tanh, sigmoid$\}$      \\ 
    & optimiser type $h_2$  & $\{$SGD, Adam, RMSprop, AdaGrad$\}$ \\ 
    & suggested dropout value $h_3$   & $\{0.001, 0.005, 0.01, 0.05, 0.1, 0.5 \}$ \\
    & learning rate $x_1$    & $10^{[-5,-1]}$    \\
    & number of neurons $x_2$    & $2^{[4,7]}$    \\
    & aleatoric variance $x_3$    & $[0.2,0.8]$    \\    
    \bottomrule
\end{tabular}
\end{table*}

We defined three real-world tasks of tuning the hyperparameters for ML algorithms: \textit{SVM-Boston}, \textit{XG-MNIST}, \textit{NAS-CIFAR10} and \textit{NN-Yacht} . 

\paragraph{\textit{SVM-Boston}} outputs the negative mean square error of support vector machine (SVM) for regression on the test set of Boston housing dataset. We use the Nu Support Vector regression algorithm in the scikit-learn package \cite{scikit-learn} and use a train/test split of $7:3$.

\paragraph{\textit{XG-MNIST}} returns classification accuracy of a XGBoost algorithm \cite{chen2016xgboost} on the testing set of the MNIST dataset. We use the $xgboost$ package and adopt a stratified train/test split of $7:3$.

\paragraph{\textit{NAS-CIFAR10}} performs the architecture search on convolutional neural network topology for CIFAR10 classification. We use the public architecture dataset, NAS-Bench-101 \cite{ying2019bench} \footnote{Code and data are available at \url{https://github.com/google-research/nasbench}}, which contains the precomputed training, validation, and test accuracies of $423,624$ unique neural networks on CIFAR10 after training for 108 epochs. All these networks are exhaustively generated from a graph-based search space. The search space comprises a 7-node directed acyclic graph(DAG) with the first node being the input and the last node being the output. The 5 intermediate nodes can perform one of the following 3 operations: 3x3 convolution, 1x1 convolution, and 3x3 max-pooling (i.e. 5 categorical variables, each with 3 categorical choices). There are 21 possible edges in the DAG but any valid architecture is limited to a maximum of 9 edges. We follow the encoding scheme in \cite{ying2019bench}, which defines a probability value $x_i \in [0,1]$ for each possible edge $i$ and defines an integer parameter $x_22 \in [0, 9]$. An architecture is generated by activating the $x_22$ edges with the highest probability. The design of the search space turns the neural architecture search into an optimisation problem involving multiple categorical variables and continuous variables. This dataset enables us to quickly evaluate the architectures proposed by BO algorithms by looking up the dataset and compare the search strategies without the need for huge computing resources. 

\paragraph{\textit{NN-Yacht}} returns the negative log likelihood of a one-hidden-layer neural network regressor on the test set of Yacht hydrodynamics dataset. We follow the MC Dropout implementation and the random train/test split on the dataset proposed in \cite{Gal2016Dropout}\footnote{Code and data are available at \url{https://github.com/yaringal/DropoutUncertaintyExps}}. The simple neural network is trained on mean squared error objective for 20 epochs with a batch size of $128$. We run $100$ stochastic forward passes in the testing stage to approximate the predictive mean and variance. The results of CoCaBO against other methods in the batch setting ($b=4)$ for this task is shown in Figure \ref{fig:bnn_yacht} and against, CoCaBOs outperform other methods. 

The hyperparameters over which we optimise for each above-mentioned ML task are summarised in Table \ref{table:real_problem_inputs_details}. One point to note is that we present the unnormalised range for the continuous inputs in Table \ref{table:real_problem_inputs_details} but normalise all continuous inputs to $[-1, 1]$ for optimisation in our experiments. All the remaining hyperparameters are set to their default values. 

\bibliographystyle{icml}
\bibliography{mabbobib}